\documentclass{article}
\usepackage{arxiv}
\usepackage[utf8]{inputenc} 
\usepackage[T1]{fontenc}    
\usepackage{hyperref}       
\usepackage{url}            
\usepackage{booktabs}       
\usepackage{amsfonts}       
\usepackage{nicefrac}       
\usepackage{microtype}      
\usepackage{lipsum}
\usepackage{fancyhdr}       
\usepackage{comment} 
\usepackage{longtable, makecell,multirow}

\usepackage{xcolor}

\usepackage[colorinlistoftodos]{todonotes}

\usepackage{color}
\definecolor{dkgreen}{rgb}{0,0.6,0}
\definecolor{gray}{rgb}{0.5,0.5,0.5}
\definecolor{mauve}{rgb}{0.58,0,0.82}
\newcolumntype{R}[1]{>{\raggedleft\arraybackslash }p{#1}}
\newcolumntype{L}[1]{>{\raggedright\arraybackslash }p{#1}}
\newcolumntype{C}[1]{>{\centering\arraybackslash }p{#1}}

\title{How to estimate carbon footprint when training deep learning models? A guide and review}

\author{
  Lucía Bouza Heguerte,\\
  Université Paris Cité, CNRS, MAP5 UMR 8145, 75006, Paris, France \\
   \AND
  Aurélie Bugeau \\
  Univ. Bordeaux, Bordeaux INP, CNRS, LaBRI, Talence, France\\
  IUF\\
  \texttt{aurelie.bugeau@u-bordeaux.fr} \\
   \AND
   Loïc Lannelongue\\
        Cambridge Baker Systems Genomics Initiative, Department of Public Health and Primary Care\\
        British Heart Foundation Cardiovascular Epidemiology Unit, Department of Public Health and Primary Care\\
        Victor Phillip Dahdaleh Heart and Lung Research Institute\\
        Health Data Research UK Cambridge, Wellcome Genome Campus\\
        University of Cambridge, Cambridge, United Kingdom
}
\date{}

\begin{document}
\maketitle

\begin{abstract}
Machine learning and deep learning models have become essential in the recent fast development of artificial intelligence in many sectors of the society. It is now widely acknowledge that the development of these models has an environmental cost that has been analyzed in many studies. Several online and software tools have been developed to track energy consumption while training machine learning models. In this paper, we propose a comprehensive introduction and comparison of these tools for AI  practitioners  wishing  to  start  estimating the  environmental  impact of their work. We review the specific vocabulary, the technical requirements for each tool. We compare the energy consumption estimated by each tool on two deep neural networks for image processing and on different types of servers. From these experiments, we provide some advice for better choosing the right tool and infrastructure. 
\end{abstract}

\section{Introduction}

Deep learning has been widely used in every sector of the society for a few years. A search of Scopus shows that it went from about 1,350 research papers in  2015 to more than 85,000 in 2022. Results obtained in every domain are impressive, and AI is a promising tool for tackling environmental challenges in particular \cite{rolnick_tackling_2019, vinuesa_role_2020, kar_how_2022}. But it is also now widely documented that training and deploying deep learning projects has an impact on the environment~\cite{strubell_energy_2019,gupta_chasing_2022, gupta2020secure,  ligozat2022unraveling, kaack_aligning_2021, lannelongue2023carbon, bannour2021evaluating, thompson2020computational, strubell22, Henderson20}. These studies have assessed energy consumption and corresponding amount of greenhouse gas emissions (in CO2 equivalent, denoted as CO2eq) from computer calculations when training a deep learning program, and showed that recent large language models can be responsible for hundreds of tonnes of CO2eq~\cite{luccioni2022estimating}, whereas, for context, a limit of 2 tCO2eq/person/year is what is needed to keep global warming under 1.5$^{\circ}$C~\cite{Arias_IPCC_6}. Some studies have also compared existing estimation tools 
~\cite{bannour2021evaluating,lannelongue2023carbon, jay:hal-04030223}. 

Despite these many studies, when AI practitioners wish to start estimating their environmental impact, they may face several difficulties. Depending on their backgrounds, it might be difficult for them to get used to the hardware-related vocabulary, know how to use the estimation tools (locally or on servers), and determine which tool is best suited for their current use-case. This document aims to address these and ease the process of energy consumption measurement for AI practitioners.
It can be used as a guide to measure the energy consumption and associated greenhouse gas emission when training deep learning algorithms and although what will be explained can be applied to other types of algorithms and other infrastructures, we will focus on training deep-learning models in different types of infrastructures.

In this context, this document makes the following contributions:
\begin{itemize}
\item We review existing tools for measuring or estimating the energy consumption of computations, and explain the specific notions that are not always known by AI practitioners. It goes further than previous surveys~\cite{bannour2021evaluating,lannelongue2023carbon, jay:hal-04030223} in providing details about what is measured by each tool and on which infrastructure they can be used, the measurement process, how usage factor is being used, default values, and the source of information that are used. These information are crucial to correctly interpreting the data obtained. 
\item We test and compare these different approaches using wattmeters to assess their accuracy. We also quantify the energy consumption of the estimation tools themselves.
\item We run a range of experiments to analyze the influence of key hyperparameters such as batch size, data load, checkpoints and epochs. These lead to a set of recommendations on how and when to use these tools depending on the infrastructure available to train the models. For instance, we show that it seems possible to only measure part of training and extrapolate to avoid the small extra consumption from energy measurement. We also show that batch size can influence energy consumption. The recommendations complete previous works that intended to make machine learning researchers better understand their carbon impact and to take steps to mitigate it~\cite{ligozat:hal-03376391, strubell22}.

\end{itemize}

The seven different tools that we study are:    \href{https://www.green-algorithms.org}{Green-Algorithms} \cite{GreenAlgorithms}~(GA), \href{https://codecarbon.io}{CodeCarbon} \cite{CodeCarbon} (CC (P) for process, CC (M) for machine),  \href{https://github.com/sb-ai-lab/Eco2AI}{Eco2AI} \cite{Eco2AI} (E2 (P) for process, E2 (M) for machine), \href{https://github.com/lfwa/carbontracker/tree/master}{CarbonTracker} \cite{CarbonTracker} (CT), \href{https://github.com/Breakend/experiment-impact-tracker}{Experiment-Impact-Tracker} \cite{ExperimentImpactTracker} (EIT), \href{https://mlco2.github.io/impact/}{MLCO2} \cite{MLCO2} and \href{https://github.com/epfl-iglobalhealth/cumulator}{Cumulator} \cite{Cumulator} (CMLTRs). 

We use the following infrastructures, all located in France, for training models: Labri servers (institutional server), MAP5 servers (institutional server), Grid5000 distributed cluster and personal computers. Mention will also be made of the Google Colab environment.

In the Labri servers, personal computer and in Grid5000 there are wattmeters (WM), 
which can provide real information on the consumption of energy of the infrastructure in a given period. \\ 

\begin{figure*}[h!]
\centering
\includegraphics[width=0.9\textwidth]{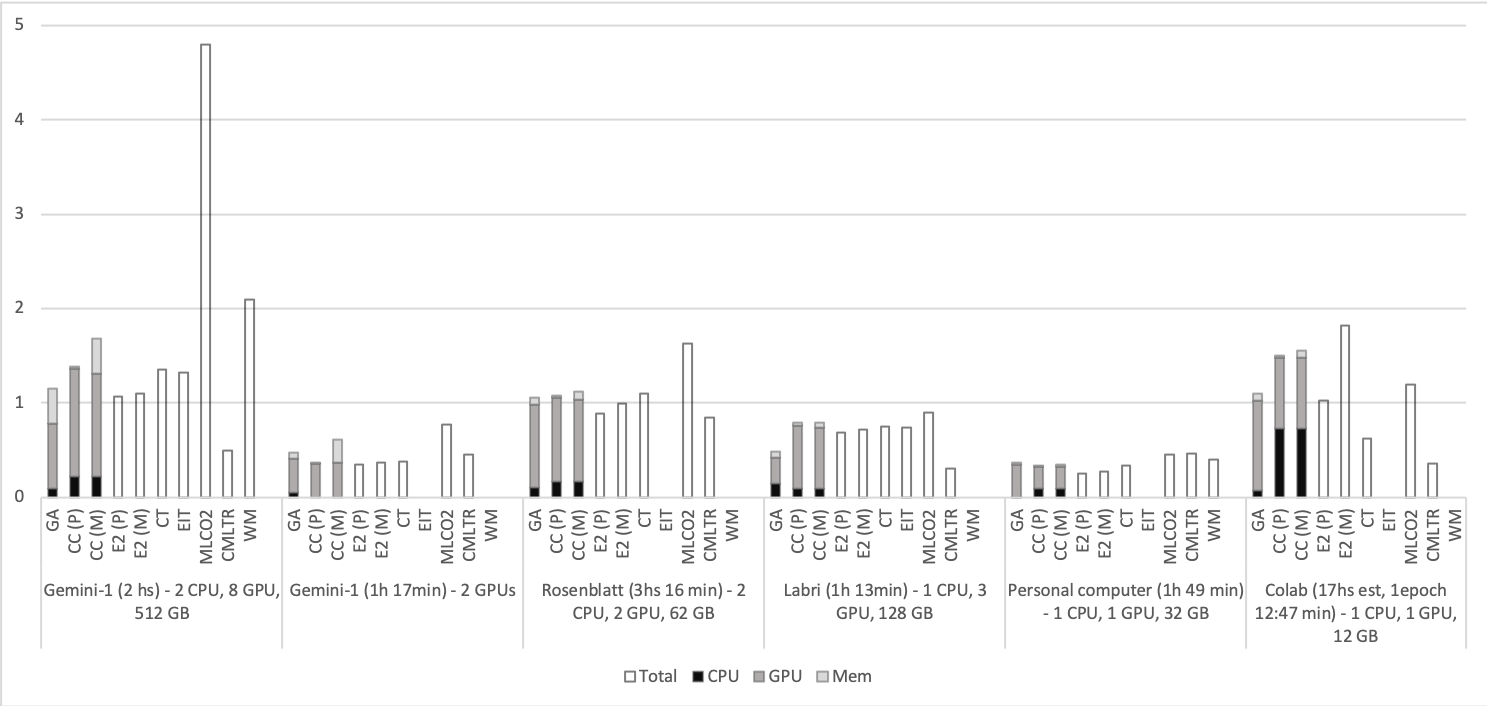}
\caption{Energy consumption in kWh of the different methods over the 5 different infrastructures for the second experiment. For the tools that do not provide detail for CPU/GPU/Memory consumption, the total energy reported is plotted. }
\label{fig:Expe2}
\end{figure*}

We focus on two machine learning experiments, both for image processing. In the first one, a small neural network is trained for digit classification on the MNIST dataset \cite{deng2012mnist}. This experiment is short, approximately 1 minute. In the second, a DNCNN network is trained for noisy image denoising. The training is carried out with the Imagenet validation dataset \cite{deng2009imagenet}. This experiment is longer, approximately 2 h.\\
\begin{figure*}[h!]
\centering
\includegraphics[width=0.9\textwidth]{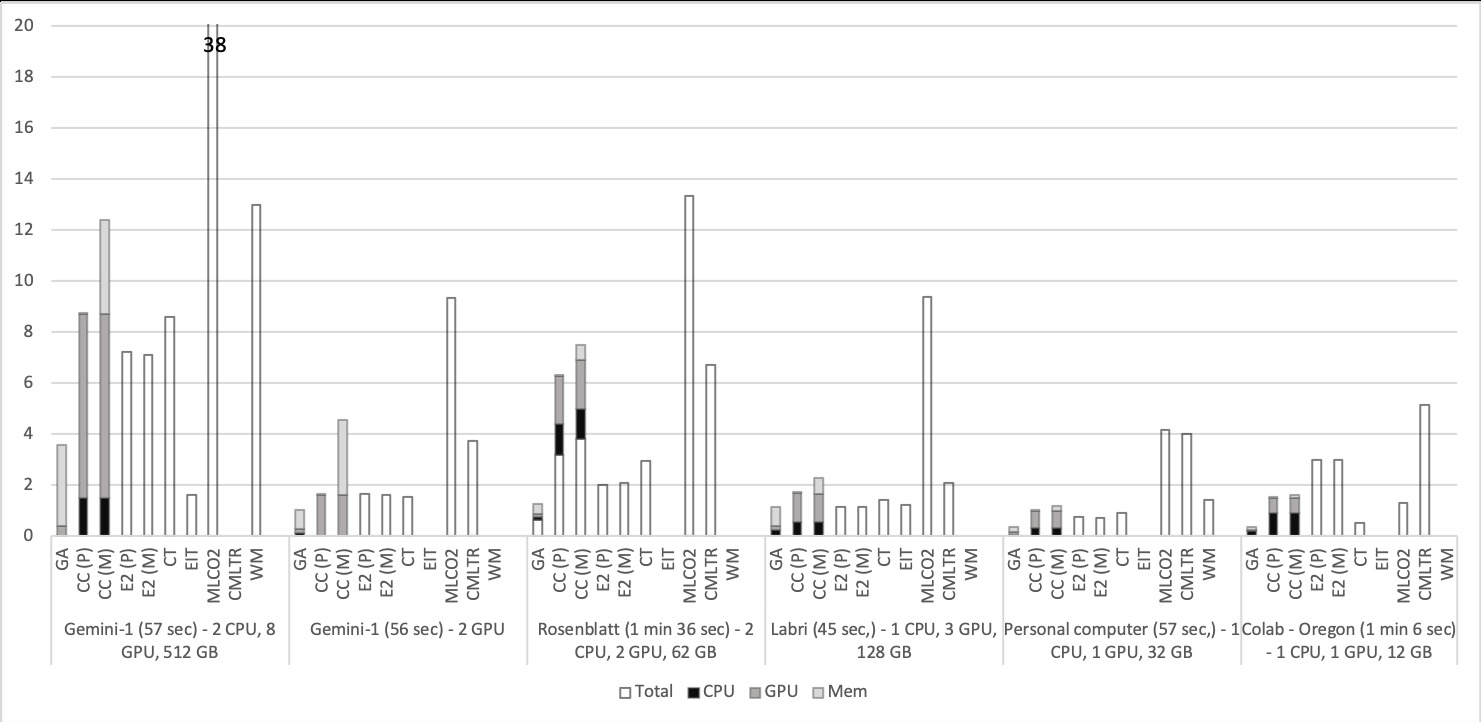}
\caption{Energy consumption in Wh of the different methods over the 5 different infrastructures for the first experiment. For the tools that do not provide detail for CPU/GPU/Memory consumption, the total energy reported is plotted.  }
\label{fig:Expe1}
\end{figure*}
Figures~\ref{fig:Expe1} and~\ref{fig:Expe2} summarize the energy consumption of the different tools in the five tested infrastructure. 
As we detail in this guide, the high variability comes from the different goals of the different tools, some estimate the power consumption of the entire machine while others focus on a particular process. The idle power consumption is also accounted for differently, alongside usage factors, CPUs vs GPUs etc.

The document is organized as follows. Users already familiar with carbon footprint estimation may directly jump to section \ref{experes} for the results.  Section \ref{related} reviews previous publications in this field.  Section \ref{sec:tools} details the specificities of each tool: energy consumption of each hardware components and their communications, power usage effectiveness and emission intensity. Section \ref{Infrastructure} details the type of infrastructures that are typically used to train AI models and what tools can be used for each.
Section \ref{experes} presents the experimental setup and an analysis of the results. Discussions on the results presented and recommendations on when and how to estimate all environmental impacts end this guide in section \ref{discussions}. Finally, errors reported and found in the tools are added in the Appendix.
\section{Related works}\label{related}
Only recently estimation tools have been made available and consequently, few studies have compared and analyzed existing strategies for measuring energy consumption of deep learning projects. 

The authors of \cite{bannour2021evaluating} reviewed six tools (CarbonTracker, Experiment-impact tracker, Green-Algorithms, MLCO2, energy usage and Cumulator) that are available to measure energy use and CO2eq emissions in the context of natural language processing. They compared the tools according to publication details, technical criteria (availability, online, easy-of-use, documentation etc.), configuration criteria (specification of carbon intensity, PUE, install dependent, etc.) and functional criteria (idle power and communication between hardware). The authors observed a two-fold variation in estimates between tools and concluded that further studies are needed to better understand these tools and estimate broader impacts. 

In the same line of research, the authors of \cite{jay:hal-04030223} compared some tools on server nodes, not all specifically designed for deep learning and therefore not all integrating GPUs. They categorized tools between external and internal node sensors, power profiling software, energy measurement software packages and online energy calculators. They looked at publication year, environment criteria (hardware compatibility, virtualization, etc.), functional criteria (hardware compatibility, software power model, sampling frequency, reporting and profiling), and user-friendliness. They tested each tool on the same server nodes and compared them with external power meters. The authors drew some recommendations from this study: to monitor power consumption in real time, it is better to use power profiling software, but they do not measure GPUs consumption; relationship between energy measurement software tools and power meter is not constant, so software tools are not perfectly accurate. 

Finally, \cite{lannelongue2023carbon} provided general guidelines about the strength and weaknesses of different types of estimation tools, namely online calculators, embedded packages and server-side tools. The criteria that are discussed are compatibility with any hardware, any programming language, research field, some ease of use criteria and scalability with number of jobs and long periods of time.

The different tools discussed above focus on energy consumption during the training phase of AI models, which only constitute part of the broader environmental impacts of AI \cite{gupta_chasing_2022, gupta2020secure,  ligozat2022unraveling, kaack_aligning_2021, lannelongue2023carbon}. In this context, the authors of \cite{luccioni2022estimating} later included embodied impacts as well as emissions from static infrastructure and deployment when studying BLOOM, a large language model.

\section{Estimating greenhouse gas emissions}\label{sec:tools}

This section explains how tools measure or estimate energy consumption and CO2eq emissions, from Python libraries integrated into the code (referred to as software tools), to web forms and physical watt measurement devices connected to the infrastructure used. Some of these tools also have a server-side version, to be used in HPC clusters and thus be able to collect information more easily to estimate energy consumption. Online tools and server-side tools can be used without modifying the code, and are independent of the programming language used. 
Python libraries can only be used in Python code but enable measurements of the consumption of different parts of the programs. Watt-measurement enables measuring the consumption of the whole node but are not always available and can not isolate a paritcular process.  
Each tool has its own way of estimating the consumption of each component. A summary of the characteristics is shown in the table \ref{tableTools}.

The most power consuming devices on a personal machine or a server are the GPUs (if present), CPUs, and memory. There are other resources, such as storage or the network, that are generally not considered in software measurements, since they do not provide a significant load over the duration of an AI task. Indeed, in regular use, storage is typically solicited far less than memory and is mainly used as a more permanent record of the data, independently of the task~\cite{GreenAlgorithms}.

When the machine is in a data center, energy usage of all equipment that are necessary to power, cool and maintain the datacenter should be measured as they may account for an important amount of energy consumption. This is done using the efficiency coefficient of the data center called power usage effectiveness (PUE).

\subsection{Energy consumption of each component}

In this section, we will see the different strategies used by the tools to estimate the energy consumed by the different resources and estimate the consumption of the processes.  Green-Algorithms and CodeCarbon are the only Python tools that report the estimate of consumed or emissions, discriminated by each component: memory, CPU and GPU.  

A transversal concept to all resources is the usage factor. The usage factor of a resource refers to the percentage of use that can be assigned to the process being measured. For example, if the CPU power is estimated to 2W, but the CPU usage factor of the process was 50\%, then the consumption of a one hour process is assumed to be 1 kWh. If the usage factor is unknown, then 100\% of the use of the resource is being assigned to the process, when in fact there may be other processes also using said resource.

During the measured period, some tools query sensors or perform calculations to estimate power consumption. Note that Lower  measurements frequency mean fewer measurements that may lead to more approximate results. By default, CodeCarbon performs these measurements every 15 seconds. Eco2AI, CarbonTracker and Experiment-Impact-Tracker take measurements every 10 seconds. Cumulator does not query sensors or intermediate measurements to estimate energy consumption. 

\subsubsection{Energy consumed by CPU\\}

There are two methods used in the tools to estimate energy consumed by CPUs: using CPU thermal design power (TDP) provided by the manufacturer, or using software integrated tools (RAPL files or Power Gadget). Appendix \ref{AppTDP} provides explanations of these two methodologies. Note hat software integrated tools may require privilege permissions as summarized in section \ref{AccessInformationAndResources}. We review in table~\ref{tab:CPUs} how CPU power consumption is measured in AI measurement tools.

\begin{longtable}{p{0.15\textwidth}p{0.8\textwidth}}
    \caption{Estimation of energy consumption for CPUs}\label{tab:CPUs}\\ 

         \hline
        \multicolumn{2}{l}{  \textbf{Green-Algorithms}}\\
        Energy & uses the model of CPU provided by the user to pull the corresponding TDP from a database, or the user can input the TDP manually. If TDP is unknown, GA uses an average of 12W per core, but the paper does not explain this value. In this model, a core power usage is assumed to be equal to the TDP divided by the number of cores (if a chip has 2 cores and a TDP of 50W, then the TDP per core is 25). \\
        Usage factor & uses usage factors if known, and assumes 100\% usage if not.\\
        \hline
        \multicolumn{2}{l}{  \textbf{CodeCarbon}}\\
        Energy & uses RAPL files or Power Gadget to report CPU energy consumption (only for INTEL CPUs  with root access). The consumption reported by RAPL files or Power Gadget represents the consumption of the whole machine, and not only the process. If CodeCarbon cannot find the software to track the CPUs, then the tool uses the model of CPU to search in a list the corresponding TDP. If the model is unknown, it uses a TDP of 85W. The authors do no specify where is this value taken from.\\
        Usage factor & Not computed when using RAPL files or Power Gadget \\
        &When TDP is used, CodeCarbon assumes that the average usage factor is 50\% but this value is not explained and seems arbitrary.\\
         \hline
        \multicolumn{2}{l}{  \textbf{Eco2AI}}\\
        Energy & uses the model of the CPU to search in a list the corresponding TDP. If TDP is unknown, it uses an average of 100W~\cite{maevsky2017evaluating}. \\
         Usage factor & uses \textit{os} and \textit{psutil} python modules to determine usage factor if the tracking mode \textit{current} is set (default).\\
         \hline
        \multicolumn{2}{l}{  \textbf{CarbonTracker}}\\
        Energy & uses RAPL files to report CPU energy consumption (only for INTEL CPUs with root access). Without access to the RAPL files, the tool will not measure CPU. CarbonTracker will work only if it can measure at least one component (CPU or NVIDIA GPU).\\
        Usage factor & not computed. The power consumption values of the RAPL files are global to the whole machine.
\\
         \hline
         
        \multicolumn{2}{l}{  \textbf{Experiment-Impact-Tracker (EIT)}}\\
        Energy & uses RAPL files to report CPU energy consumption (only for INTEL CPUs with root access and Linux operating system)\\
        Usage factor & uses \textit{psutil} python module to determine usage factor
\\
         \hline
   
         \textbf{MLCO2} &does not measure CPU utilization. \\
  
         \hline
        \multicolumn{2}{l}{  \textbf{Cumulator}}\\
        Energy & It is not possible to measure GPU and CPU components at the same time but Cumulator measures CPU utilization by default. It uses the model of CPU to search in a list for the corresponding TDP. If TDP is unknown, it uses an average of 250W. This value is the one of Nvidia GeForce GTX Titan X, which is the GPU model in the IC cluster of the EPFL Machine Learning and Optimization Laboratory (MLO). It considers just one CPU.\\
        Usage factor & does not use usage factor.
\\
         \hline

\end{longtable}

file:///Users/bugeau/Recherche/Impact/IA/ComparaisonOutils/Guide to measure carbon footprint when training deep learning models in France/bibliography.bib

\subsubsection{Energy consumed by GPU\\}
As with CPUs, energy consumption for GPUs are computed either from TDPs provided by manufacturers or from internal tools. The latter is done with the \textit{pynvml} library that only works for Nvidia GPUs.  We review in table~\ref{tab:GPUs}  how GPU power consumption is measured in AI measurement tools.

\begin{longtable}{p{0.15\textwidth}p{0.8\textwidth}}
    \caption{Estimation of energy consumption for GPUs}\label{tab:GPUs}\\ 
         \hline
        \multicolumn{2}{l}{  \textbf{Green-Algorithms}}\\
        Energy & uses the model of GPU to search in a list the corresponding TDP. You can load the TDP of the GPU if the model is not listed. If TDP is unknown, it uses an average of 200W, but the paper does not explain the reason for choosing this value. \\
        Usage factor & GPUs usage factor is considered if known by the user. If not, GA considers 100\% of usage. \\
         \hline
        \multicolumn{2}{l}{  \textbf{CodeCarbon}}\\
        Energy & uses \textit{pynvml} library (only for NVIDIA GPUs). CodeCarbon does not measure  consumption of non-NVIDIA GPUs.\\
         Usage factor & not computed. The consumption reported by pynvml represents the consumption of the whole machine, and not only the process.\\
         \hline
        \multicolumn{2}{l}{  \textbf{Eco2AI}}\\
        Energy & uses \textit{pynvml} library (only for NVIDIA GPUs). Eco2AI does  not measure consumption of non-NVIDIA GPUs.\\
         Usage factor & not computed. The consumption reported by pynvml represents the consumption of the whole machine, and not only the process.\\
         \hline
        \multicolumn{2}{l}{  \textbf{CarbonTracker}}\\
        Energy &uses \textit{pynvml} library (only for NVIDIA GPUs). CarbonTracker does  not measure consumption of non-NVIDIA GPUs.\\
        Usage factor & not computed. The consumption reported by pynvml represents the consumption of the whole machine, and not only the process.\\
         \hline
        \multicolumn{2}{l}{  \textbf{EIT}}\\
        Energy & uses \textit{nvidia-smi} command line (only for NVIDIA GPUs). EIT does  not measure consumption of non-NVIDIA GPUs.\\
        Usage factor & uses \texttt{Popen} to open a thread, execute the command \texttt{nvidia-smi -q -x}, get the output in a xml, and parse it to get the usage factor of the GPU.
\\
         \hline
        \multicolumn{2}{l}{  \textbf{MLCO2}}\\
        Energy &  uses the model of GPU to search in a list the corresponding TDP. It is not possible to load the TDP of the GPU if the model is not listed. In this case, it is necessary to do a pull request to add the value. It is not possible to choose the quantity of GPUs.\\
        Usage factor & does not use usage factor. The GPU is considered at maximum load and this load is assumed to correspond to the measured process.
\\
         \hline
        \multicolumn{2}{l}{  \textbf{Cumulator}}\\
        Energy & uses the model of GPU to search in a list the corresponding TDP. If TDP is unknown, it uses an average of 250W.  It considers just one GPU. \\
        Usage factor &does not use usage factor. The GPU is considered at maximum load and this load is assumed to correspond to the measured process.
\\
         \hline

\end{longtable}

\subsubsection{Energy consumed by memory\\}

According to \cite{Hodak2019TowardsPE} GPUs are responsible for around 70\% of power consumption, CPU for 15\%, and RAM for 10\%.

Some tools like Green-Algorithms consider that power consumption of RAM depends strongly on the available memory, independently of the memory consumed \cite{karyakin2017,Guo_Yu_2022}, while other tools like Eco2AI considers that it depends on the allocated memory by the process \cite{maevsky2017evaluating}.
 We review in table~\ref{tab:Mems}  how memory power consumption is measured in AI measurement tools.

\begin{longtable}{p{0.25\textwidth}p{0.7\textwidth}}
    \caption{Estimation of energy consumption for memory}\label{tab:Mems}\\ 
         \hline
        {  \textbf{Green-Algorithms}} & Energy consumption by memory is 0.3725W/GB of memory available (If we have all the server memory available, it will account for all the server memory. If we are in an HPC cluster, it will account only for the amount of memory requested, regardless of how much the process consumes). The value 0.3725 was obtained experimentally \footnote{Source: \href{https://www.tomshardware.com/reviews/intel-core-i7-5960x-haswell-e-cpu,3918-13.html}{www.tomshardware.com}}. \\
         \hline
        {  \textbf{CodeCarbon}} & Energy consumption by memory is 0.375W/GB of memory used\footnote{Source: \href{https://www.crucial.com/support/articles-faq-memory/how-much-power-does-memory-use}{Crucial}}. If tracking mode is “process”, the memory used by the process is measured via \textit{psutil}.\\
         \hline
       {  \textbf{Eco2AI}}&Energy consumption of memory is 0.375W/GB of memory used \cite{maevsky2017evaluating}. Memory used by the process is measured via \textit{psutil}.\\
         \hline
        {  \textbf{CarbonTracker}} & uses RAPL files to report memory energy consumption. It measures the total energy of memory available, not only the one used by the process. Without access to the RAPL files, the tool will not measure memory energy consumption.\\
         \hline
        {  \textbf{EIT}} & uses RAPL files or Power Gadget to report memory energy consumption. Memory used by the process is measured via \textit{psutil} considering memory used exclusively by the process and the shared memory between processes (weighted by the number of processes). Without access to the RAPL files or Power Gadget, the tool cannot be used.
\\
         \hline
       {  \textbf{MLCO2}} &does not measure memory.\\
         \hline
      {  \textbf{Cumulator}}&does not measure memory.\\
         \hline

\end{longtable}

\subsubsection{Energy consumed by communications \\}
\label{sec:com}
In ICT (Information and Communication Technology), communications refer to the exchange of information or data between two or more nodes. Nodes can be any device that is connected to a network, including computers, routers, servers, and even mobile devices. Machine Learning algorithms typically involve the exchange of data between nodes at various stages, such as during data generation, during training (parameter updates across different nodes in the network), or while the model is in production. 

The only tool that estimates the cost of communications is  Cumulator. Each time the model sends a data file to another node of the network, Cumulator records the size of the file which is communicated. The cost of communication relies on the "1byte model" of the Shift Project \cite{lean-ict}. The value from 2017 is $6.894 \times 10^{-11}$ kWh/B. 

\subsection{PUE}

Power Usage Efficiency is the efficiency coefficient of the data center. If PUE is not given, we recommend considering the 2022 average value of 1.55~\cite{uptimeinstitute2022}. For personal computers, PUE=1 as there are no other large devices consuming power. 
We review in table~\ref{tab:PUEs} the PUEs used by each tools. All except Cumulator report the total energy consumed, including PUE. To calculate this value for Cumulator, we can divide  the reported value of greenhouse gas emissions ($GHG$) by the emission intensity ($EI$) of servers location: $Energy = GHG/EI$. 
Note that for the purpose of comparing reported energy consumption between tools, PUE is not taken into account, since each tool uses a different value.

\begin{longtable}{p{0.25\textwidth}p{0.7\textwidth}}
    \caption{PUE values used in the different tools}\label{tab:PUEs}\\ 
         \hline
        {  \textbf{Green-Algorithms}} &  configurable. The default value is 1.67 (2019)~\cite{lawrence2019}.\\
         \hline
        {  \textbf{CodeCarbon}} &not taken into consideration, except for cloud providers.\\
         \hline
       {  \textbf{Eco2AI}}& configurable. The default value is 1.\\
         \hline
        {  \textbf{CarbonTracker}} & configurable. Although the paper indicates that the 2020 PUE (1.58) is used, the 2022 PUE (1.55) is used in the code \cite{uptimeinstitute2022}.\\
         \hline
        {  \textbf{EIT}} & configurable. The default value is 1.58 (2020)~\cite{lawrence2020}.
\\
         \hline
       {  \textbf{MLCO2}} &not taken into consideration.\\
         \hline
      {  \textbf{Cumulator}}&not taken into consideration.\\
         \hline

\end{longtable}

\subsection{Carbon emission and emission intensity}

The origin of the energy used is key when determining greenhouse gas emissions from electricity production. To carry out the calculation, the average emission intensity (or carbon intensity) of the country or region where the calculations were made is used. Countries report these values, which can then be used by the tools to calculate emissions.

It is important to mention that most of the tools do not yet take the information of carbon intensity in real time. Only CarbonTracker (for UK and Denmark) and Experiment-Impact-Tracker (for California) do it.  In most cases, average values from previous years are used. Some variables, such as the time of day of execution, or the distribution of energy sources at a given moment, are not represented, but can have an important influence on the emissions, as shown on table \ref{tab:CI}. Machine learning users could look at current and planned energy consumption of most of the countries before running their experiments, e.g. on \href{https://www.electricitymap.org}{Electricity Maps}.  In some cases, if users are running on clouds that have different geographic locations, users could choose where to run the algorithms to emit fewer GHGs.  For example, table \ref{tab:CI} presents some values at different locations for two different days.  While it can be wise to carefully choose datacenter locations, developers must keep in mind that transferring large datasets from one location to the other also has environmental impacts (section~\ref{sec:com}). Therefore, depending on the training time, it might be better to remain on the same server when training on the same large dataset. We present in table~\ref{tab:Mix} how each tool handles carbon intensity.

\begin{table}[h!]
    \caption{Daily average carbon intensity for two different days. Data taken from \href{https://www.electricitymap.org}{Electricity Maps}}
    \label{tab:CI}
    \centering
    \begin{tabular}{c|cc}
         &   March 5th 2023&  March 29th 2023\\
         \hline
       France  &  64 & 137\\
      North Sweden  & 16 & 14\\
      South Africa  & 684 & 702\\
      South Carolina - USA  & 432 & 786
    \end{tabular}
\end{table}

\begin{longtable}{L{0.25\textwidth}p{0.7\textwidth}}
    \caption{Emission intensity used in the different tools}\label{tab:Mix}\\ 
         \hline
        {  \textbf{Green-Algorithms}} & Most emission intensity data come from \href{https://www.carbonfootprint.com}{Carbon Footprint} but the tool also uses other sources like \href{https://www.electricitymap.org}{Electricity Maps}. Information is collected in the \href{https://github.com/GreenAlgorithms/green-algorithms-tool/blob/master/data/latest/CI_aggregated.csv}{CI\_aggregated.csv} file. The default value is 475 gCO2eq/kWh (world average in 2018).\\
         \hline
        {  \textbf{CodeCarbon}} & For United States and Canada, CodeCarbon uses regional data on emissions per unit of power consumed. For other countries, the tool uses the energy mix of the country, i.e. intensity data of each energy source (carbon, solar, wind, etc.), to calculate the intensity of the country. 
        The average energy mix for each country is taken from \href{https://www.globalpetrolprices.com}{Global Petrol Prices}. The information is collected in the files under \href{https://github.com/mlco2/codecarbon/tree/master/codecarbon/data}{data} folder. The sources of each data are specified in the files. The default value is 475 gCO2eq/kWh (world average in 2018).\\
         \hline
       {  \textbf{Eco2AI}}&For all countries the emission intensity calculation was made using the intensity data of each energy source (carbon, solar, wind, etc.) and the energy mix of each country. The values used for the calculations nor their sources are not explained, and only the final result of the intensity of emissions for each country is published in \href{https://github.com/sb-ai-lab/Eco2AI/blob/main/eco2ai/data/carbon_index.csv}{carbon\_index.csv}. The default value is 436.5 gCO2eq/kWh~\cite{ember2022global}.\\
         \hline
        {  \textbf{CarbonTracker}} & CarbonTracker supports the fetching of carbon intensity in real-time through external APIs. It is currently limited to Denmark and Great Britain. For Denmark they use data from \href{https://energidataservice.dk/}{Energi Data Service} and for Great Britain they use the \href{https://carbonintensity.org.uk/}{Carbon Intensity API}. For other countries, it uses fixed values available in the \href{https://github.com/lfwa/carbontracker/blob/master/carbontracker/data/carbon-intensities.csv}{carbon-intensities.csv} file. The sources are not published. The default value is 475 gCO2eq/kWh (2019).\\
         \hline
        {  \textbf{EIT}} & EIT supports the fetching of carbon intensity in real-time through external APIs. It is currently limited to California using the API of \href{http://caiso.com}{California ISO}. For other countries, it uses fixed values available in the \href{https://github.com/Breakend/experiment-impact-tracker/blob/master/experiment_impact_tracker/emissions/data/co2eq_parameters.json}{co2eq\_parameters.json} file. The sources are published and are mostly from \href{https://www.electricitymap.org}{Electricity Maps}. The default value is  301 gCO2eq/kWh (annual mean carbon intensity of all electricityMap zones).\\
         \hline
       {  \textbf{MLCO2}} &MLCO2 published the sources and contains the information of the Cloud providers in the \href{https://github.com/mlco2/impact/blob/master/data/impact.csv}{impact.csv} file. For private infrastructure, it is necessary to provide the emission intensity value, which must be obtained by user own means.\\
         \hline
      {  \textbf{Cumulator}}&The data of emission intensity is from \href{https://www.electricitymap.org}{Electricity Maps}.  
Information is collected in the \href{https://github.com/epfl-iglobalhealth/cumulator/blob/master/src/cumulator/countries_data/country_dataset_adjusted.csv}{country\_dataset\_adjusted.csv} file. The default value is  447 gCO2eq/kWh (average carbon intensity value in gCO2eq/kWh in the EU in 2018 \cite{MORO20185}).\\
         \hline

\end{longtable}

\subsection{Measuring whole equipment consumption with wattmeters}

Wattmeters are physical instruments that are used to measure the active electrical energy of a certain circuit. By plugging them into the physical infrastructure, we can get the exact total consumption of the machine. With wattmeters, it is not possible to determine how much energy each component of the machine consumes, neither to discriminate consumption by process. It is also important to note that wattmeters have measurement frequencies. Different wattmeters may have different measurement frequencies and therefore different accuracies depending on the duration of processes.

\subsection{Errors reported and found in the tools}

Some tools had to be modified to be used, as they had bugs not yet fixed by the authors. The modifications we had to make can be found in~\ref{bug}.

\subsection{Summary of the characteristics of existing tools}

In addition to the tables presented in ~\cite{bannour2021evaluating} and \cite{jay:hal-04030223}, we summarize in table~\ref{tableTools} what is configurable and what are default values for each component, and add details on usage factor.

\begin{table}[!h]
    \caption{Summary of the characteristics of the energy and CO2eq measurement tools. Wattmeters are not included in the table. \label{tableTools}}
   \resizebox{\textwidth}{!}{%

\setlength{\tabcolsep}{0.2cm} 
    \begin{tabular}{ L{5cm}|C{3cm} C{3cm} C{3cm} C{3cm} C{3cm}C{3cm}C{3cm}}

        ~ & \textbf{Green-Algorithms} & \textbf{CodeCarbon} & \textbf{Eco2AI} & \textbf{CarbonTracker} & \textbf{EIT} & \textbf{MLCO2} & \textbf{Cumulator} \\ \hline
        \multicolumn{8}{l}{  \textbf{General Information}}\\\hline
        1. Type of tool & Online calculator and Server-side tool & Embedded package & Embedded package & Embedded package & Embedded package & Online calculator & Embedded package \\ \hline
        2. Embodied emissions & no & no & no & no & no & no & no \\ \hline
        3. Static (idle) emissions w/o runs & no & no & no & no & no & no & no \\  \hline
        
        4. Process/machine estimation & process & both & both  & machine & process & machine & machine \\  \hline
        5. Measurement frequency (sec) & - & 15 & 10 & 10 & 10 & - & - \\  \hline
        
        \multicolumn{8}{l}{  \textbf{Energy Consumption CPU}}\\\hline
        1. Measured & yes & yes & yes & yes & yes & no & yes (if chosen) \\ \hline
        2. Use Model of CPU & yes & yes (if no tracking tool) & yes & no & no & - & yes \\ \hline
        3. Use RAPL files or Power Gadget & no & yes & no & yes (RAPL files) & yes & - & no \\ \hline
        4. Default TDP & 12 (normalized by core) & 85 & 100 & - & - & - & 250 \\ \hline
        5. Usage Factor considered & yes & 50\% (if default TDP used) & yes & no & yes & - & no \\ \hline
        6. Tool for usage factor & - & - & psutil & - & psutil & - & - \\ \hline
        
        \multicolumn{8}{l}{  \textbf{Energy Consumption GPU}}\\\hline
        1. Measured & yes & yes & yes & yes & yes & yes & yes (if chosen) \\ \hline
        2. Use Model of GPU & yes & no & no & no & no & yes & yes \\ \hline
        3. Default TDP & 200 & no & no & no & no & no & 250 \\ \hline
        4. Tool to get power & - & pynvml & pynvml & pynvml & nvidia-smi & - & - \\ \hline
        5. Usage Factor considered & yes & no & no & no & yes & no & no \\ \hline
        6. Tool for usage factor & - & - & - & - & nvidia-smi & - & - \\ \hline
        7. Only Nvidia GPUs & no & yes & yes & yes & yes & no & no \\ \hline
        
        \multicolumn{8}{l}{  \textbf{Energy Consumption Memory}}\\ \hline
        1. Measured & yes & yes & yes & yes & yes & no & no \\ \hline
        2. Source of information & - & system & system & RAPL files & RAPL files & - & - \\ \hline
        3. Usage Factor considered & no & yes (if tracking mode) & yes & no  & yes   & - & - \\ \hline
        4. Tool for usage factor & - & psutil & psutil & - & psutil & - & - \\ \hline
        5. Formula & 0.3725 W/GB & 0.375 W/GB & 0.375 W/GB & - & - & - & - \\ \hline
        
        \multicolumn{8}{l}{  \textbf{Emission intensity}}\\ \hline
        1. Default E.I value & 475 & 475 & 436.5 & 475 & 301 & - & 447 \\ \hline
        2. Real time & no & no & no & yes (just UK and Denmark) & yes (just California) & no & no \\ \hline
        
        \multicolumn{8}{l}{  \textbf{PUE}}\\ \hline
        1. PUE considered  & yes & yes (just cloud) & yes & yes & yes & no & no \\ \hline
        2. PUE configurable  & yes  & no & yes & no & yes & - & - \\ \hline
        3. Default PUE value  & 1.67  & - & 1 & 1.58 & 1.58 & - & - \\
        
        \hline
        \multicolumn{8}{l}{  \textbf{Errors}}\\ \hline
        1. Need code modification  & -  & - & - & yes (with Python 3.10) & yes & - & yes \\
        \hline
    \end{tabular} }

\end{table}

\section{Infrastructure}\label{Infrastructure}

Depending on the infrastructure, users will have access to different resources, which restricts the list of tools that can be used. The most commonly used infrastructures for machine learning are physical or virtual servers, virtualized environments in the cloud, supercomputers or personal computers. Table \ref{tableTools2} summarizes the tools' requirements and hardware compatibility.

\begin{table}[!h]
    \caption{Requirements to run the tools.
    \label{tableTools2}}
    \resizebox{\textwidth}{!}{%

\setlength{\tabcolsep}{0.2cm} 
    \begin{tabular}{ L{5cm}|C{3cm} C{3cm} C{3cm} C{3.5cm} C{2.5cm}C{3cm}C{3cm}}
    \hline
        ~ & \textbf{Green-Algorithms} & \textbf{CodeCarbon} & \textbf{Eco2AI} & \textbf{CarbonTracker} & \textbf{EIT} & \textbf{MLCO2} & \textbf{Cumulator} \\ \hline
        
        \multicolumn{8}{l}{  \textbf{Requirements}}\\
        \hline
        1. Operating System & - & - & - & Linux (if no-NVIDIA GPU) & - & - & - \\ \hline
        2. Access to RAPL files & no & no & no & yes (if no-NVIDIA GPU) & yes & no & no \\ \hline
        3. Power Gadget & - & no & no & - & yes & - & no \\ \hline
          \multicolumn{8}{l}{\textbf{Compatibility}}\\
        \hline
        1. Non Intel CPUs & yes & yes & yes & no & no & does not measure CPU & yes \\      \hline
        2. Non Nvidia-GPUs & yes & no & no & no & no & yes & yes  \\ \hline
    \end{tabular}%
     }
\end{table}

\subsection{Access to information and resources}\label{AccessInformationAndResources}

We explain below how each type of infrastructure handles access to hardware information.

\paragraph{Virtual environments} Some tools require knowing the available CPU model to make a better estimation. In virtual environments, the information in the \texttt{/proc/cpuinfo} file (or equivalent tools for Windows or macOS) may not be correct, and may represent some characteristics of the CPU emulated by the virtualizer. Unfortunately, from the virtual environment, there is no way for users to know exactly the real CPU that is being used for the execution. 

\paragraph{RAPL files} Some tools require read access to the RAPL files. Access to these files is restricted by default to the root user. An administrator must be asked to grant read permission to those files. Also, these files are available only if the machine has Intel CPUs, and has Linux as an operating system. A similar situation is experienced with Power Gadget: it is exclusive to Intel CPUs, and the tool need to be installed. 

\paragraph{Usage factor} Unfortunately, there is no tool that can be used with the command line that gives us the total time of the script (whole time), the CPU time and the GPU time, in order to calculate the CPU and GPU usage factor required by Green-Algorithms. However, workload managers such as SLURM commonly log this information.
One option is to take empirical and specific measurements of the use of the GPU during the execution of their algorithm using the \texttt{nvidia-smi} tool, and extrapolate that value of GPU utilization to the entire execution. It is important to note that this utilization percentage corresponds to the total utilization, and not just the utilization of the process. There could be other processes running on the available GPUs. Up to our knowledge, there is also no tool that measures GPU time for non-Nvidia GPUs.

In addition, when calculating the CPU usage factor, it is important to consider whether the infrastructure where the process is running has hyperthreading enabled. When hyperthreading is available and enabled, the hardware components of one physical core are shared between several threads. Each thread has its own set of registers, but most resources of the core are shared between the threads. Estimating the real usage factor can be difficult in this scenario. According to some studies, the maximum capacity is up to 30\% more than without hyperthreading \footnote{Source: \href{https://www.intel.com/content/www/us/en/developer/articles/tool/performance-counter-monitor.html\#cpu\_utilization}{https://www.intel.com/}}.

\paragraph{Wattmeter} Finally, using a wattmeter requires having one, and in the case of institutional infrastructure, consulting with a systems administrator to make the physical connection. It is important to note that the wattmeter will measure the consumption of the entire node, so ideally there should not be other processes running on the node, or if there are, it is key to take it into account when analyzing the value returned by the device.

\subsection{Description of the  infrastructures used for experimentation}

In this guide we have tested on resources in two French laboratories (Labri and MAP5), Grid5000, personal computers and we will also mention Google Colab. In table \ref{InfraSpecs} we detail the hardware specifications of the infrastructure used for the experiments.
\begin{table}[!h]
    \caption{Hardware specifications of infrastructure used for experiments
    \label{InfraSpecs}}
    \centering
    \resizebox{\textwidth}{!}{%
    
\setlength{\tabcolsep}{0.2cm} 
    \begin{tabular}{ L{5cm}|C{3cm} C{3cm} C{3cm} C{3cm} C{3cm}}
    \hline
        ~ & \textbf{Gemini-1 (Grid5000)} & \textbf{Rosenblatt (MAP5)} & \textbf{Server (Labri)} &  \textbf{Personal Computer} & \textbf{Colab} \\ \hline
        \textbf{Operating System} & Linux & Linux & Linux & Linux & Linux \\ \hline
        
        \multicolumn{6}{l}{  \textbf{CPU}}\\ \hline
        
        1. Quantity & 2 & 2 & 1 & 1 & 1  \\ \hline
        2. Model & Intel Xeon E5-2698 v4 & Intel Xeon E5-2609 v4 & Intel Core i9-7940X CPU @ 3.10GHz & AMD Ryzen 5 2600 Six-Core Processor & (VE) Intel Xeon CPU @ 2.20GHz\\ \hline
        3. TDP & 135W& 85W& 165W& 65W& Unknown \\ \hline
        
        \multicolumn{6}{l}{  \textbf{GPU}}\\ \hline
        1. Quantity & 8 & 2 & 3 & 1 & 1  \\ \hline
        2. Model & NVIDIA Tesla V100-SXM2-32GB & NVIDIA TITAN Xp & NVIDIA TITAN Xp & NVIDIA TITAN V & NVIDIA Tesla T4 \\ \hline
        3. TDP & 250W& 250W& 250W & 250W& 70W\\ \hline
        
        \multicolumn{6}{l}{  \textbf{Memory}}\\ \hline
        1. Quantity & 512 GB & 62 GB & 126 GB & 32 GB & 12 GB \\ \hline
        
        \multicolumn{6}{l}{  \textbf{Wattmeters}}\\ \hline
        1. Available & yes & no & yes & yes & no \\ \hline
        2. Frequency & second & - & minute & minute & - \\ \hline
    \end{tabular}%
    }
\end{table}

\subsubsection{Laboratory servers\\}
We have tested the different measuring tools in Labri (computer science laboratory of Bordeaux) and MAP5 (laboratory of applied mathematics in Paris 5 University). 
Labri has physical servers with NVIDIA GPUs, Intel CPUs and Linux operating system. We have had the possibility to experiment using Wattmeter. Access to the RAPL files is restricted to root, so the execution of the scripts need to be done by an administrator, in order to use Experiment-Impact-Tracker and CarbonTracker.\\
MAP5 has physical servers with NVIDIA GPUs, Intel CPUs and Linux operating system. Access to the RAPL files is currently available. We can test all the tools, but we do not have a Wattmeter.

\subsubsection{Super computers\\}
We experimented one super computer: Grid5000 which is a large-scale and flexible testbed for experiment-driven research in all areas of computer science, with a focus on parallel and distributed computing including Cloud, HPC and Big Data, and AI. Grid5000 cluster allows numerous configurations and is very well documented. The cluster has servers with NVIDIA GPUs, Intel CPUs, Linux operating system and access to RAPL files.  Access to Wattmeter measurements on selected nodes is possible, so that all the tools can be used. 

However, by requesting only a portion of the node, the wattmeter value, that measures the entire node, might not be really useful as other jobs can be running in the same server. Also, note that without booking the whole node, it is not possible to get user privileges so EIT cannot be used, Carbontracker will not measure CPU, and CodeCarbon will use TDP to calculate CPU consumption.

\subsubsection{Personal computers\\}

In these machines, we could install the necessary tools and enable the permissions that are required. CarbonTracker can be used if at least one of the 2 conditions is met: having Intel CPUs or NVIDIA GPUs. If neither of the two conditions is met, the tool cannot be used. The tool will measure the power consumption of the CPUs and Memory only if the CPUs are Intel, and it will measure the power consumption of the GPUs only if they are Nvidia.

If we have non-NVIDIA GPUs, we can only use Green-Algorithms, MLCO2 (if the GPU is on the list), Cumulator and CarbonTracker (if we have Intel CPUs).

If we have non-Intel CPUs, we will not be able to use Experiment-Impact-Tracker and if we have only CPUs, we will not be able to use MLCO2 either. This explains the N/A value reported in results tables.

\subsubsection{Colab\\}

Google Colab is a widely use resource, with data centers located around the world, but unfortunately the data center cannot be selected when the environment is created. The execution location can be checked with the command \texttt{curl ipinfo.io} and then using this information to determine the data center being used \footnote{\href{Google datacenters}{https://cloud.google.com/about/locations?hl=es}}.

When running a notebook, a virtual environment is generated, for which some commands are not available, users are not administrators, do not have access to RAPL files and do not know the real resources that are being used. This limits the tools that can be used. Experiment-Impact-tracker cannot be used. Green-Algorithms, CodeCarbon, Eco2AI and Cumulator can be used, assuming an average consumption. This assumption can lead to reporting values of carbon emissions that are not the correct real ones. CarbonTracker can be used, but only with GPU runtime, and will not measure energy consumption of CPU nor Memory.

\section{Experiments and results analysis}\label{experes}
We will now compare the different tools and their use in different infrastructures for image processing and analysis.
Section~\ref{expe} details the experimental settings. Then, 
in section~\ref{results} we present the results. In section~\ref{variability} we explain the high variability between the different tools, their differences with wattmeter measurements (section~\ref{res-watt}) and the impact of the infrastructure (section~\ref{res-infra}). Later, focusing more on the second experiment, we analyze the influence of the data load (section~\ref{res-data}), of the batch size  (section~\ref{res-batch}), of saving the checkpoints  (section~\ref{res-check}) and of the energy consumption of the tools themselves (section~\ref{res-tools}). Finally, we comment on additional idle consumption (section~\ref{res-idle}).

The theoretical analysis of the tools and results provides a better understanding of differences in measurement between the tools, which \cite{bannour2021evaluating} indicated was needed.

In order to also transparently acknowledge the impact of our work, we conducted an analysis using wattmeters when available and CodeCarbon when not (machine tracking) to determine the total energy consumed throughout all our experiments. The results revealed a cumulative consumption of approximately 14.5 kWh. This value includes all the runs that led to the paper. It does not include PUE.

\subsection{Experiments settings}\label{expe}
We carried out two experiments, with different characteristics, in different infrastructures.

First, we trained a manually written digit classifier on the MNIST dataset. The MNIST dataset is a collection of images of handwritten digits. Its training set has 60,000 examples, with a size of 50 MB.The classifier is implemented with a fully connected, two-layer network (an inner layer of 32 neurons, and an output layer of 10 neurons), over 5 epochs and normally takes less than a minute on different infrastructures. This experiment runs on a single GPU.

Second, we trained an image denoiser on the Imagenet validation dataset. The ImageNet dataset is a collection of images depicting diverse objects and scenes. Its validation set has 50,000 examples, with a size of 6 GB. The Denoiser is implemented with a DnCNN network \cite{RYU} over 80 epochs and takes approximately two hours to run. This experiment runs in parallel on all available GPUs. In order to measure the impact of other configurations, small variations of this experiment were also performed.

The experiments were performed using Pytorch. Since each experiment has a different configuration regarding the use of the GPUs, the choice of framework is key to enable the use of all available GPUs. PyTorch enabled multi-GPU training. This is also the case with Tensorflow, but it would have requirer additional configuration to the default installation in order to use the available GPUs.

The experiments were carried out in the infrastructures detailed in section \ref{Infrastructure}. We also ran the experiments on Gemini-1 requesting only a quarter of the resources (two GPUs, 128 GB memory and 10 cores of the 40 available). Depending on the available resources, certain tools could be used only on some infrastructures. We now discuss the main observations from our results.

\subsection{Results}\label{results}

This section presents and analyzes the results obtained for the two experiments on the different infrastructures. 
In table \ref{ResultsExp1} we present the energy consumption for the first experiment, which corresponds to the training of a manually written digit classifier. In table \ref{ResultsExp2} we present the consumption for the second experiment, which corresponds to the training of an image denoiser. 

The reported values correspond to individual runs and are not averaged values. However, multiple runs of the experiments were performed on different infrastructures to validate the consistency of these numbers. Experiment 1 was executed 3 times on Grid5000, 2 times on MAP5,  Labri and Colab. Experiment 2 was executed twice on Grid5000 and Labri.

As said before, Cumulator does not report energy consumed. The values presented in the table were not reported by Cumulator, but calculated by us from carbon footprints.

\begin{table}[!h]
    \centering
    \resizebox{\textwidth}{!}{%
    
\setlength{\tabcolsep}{0.1cm} 
    \begin{tabular}{ L{4.2cm}|C{2.2cm} C{2.2cm} C{2.2cm} C{2.2cm} C{2.2cm} C{2.2cm} C{2.2cm} C{2.2cm} C{2.2cm}C{2.2cm}}
    
    \hline
        ~ & Green-Algorithms & CodeCarbon (P) & CodeCarbon (M) & Eco2AI (P) & Eco2AI (M) & CarbonTracker & EIT & MLCO2 & Cumulator & Wattmeter \\ \hline
        
        \multicolumn{11}{l}{  \textbf{ Gemini-1 Whole node (57 sec)}}\\ \hline
        
        Tot.  Energy reported & 5.990 & 8.800 & 12.50 & 7.200 & 7.100 & 13.30 & 2.570 & 38.00 & 4.771 &  \\ \hline
        Tot.  Energy w/o PUE & 3.590 & 8.80 & 12.50 & 7.200 & 7.100 & 8.580 & 1.630 & 38.00 & 4.771 & 13.00 \\ \hline
        Energy for  CPU & 0.007 & 1.500 & 1.500 & - & - & - & - & - & - & - \\ \hline
        Energy for  GPU & 0.395 & 7.200 & 7.200 & - & - & - & - & - & - & - \\ \hline
        Energy for  Memory & 3.16 & 0.0184 & 3.700 & - & - & - & - & - & - & - \\ \hline
        Carbon emissions & 0.307 & 0.480 & 0.690 & 0.490 & 0.480 & 0.777 & 0.140 & 2.53 & 0.563 &  \\ \hline
        
        \multicolumn{11}{l}{  \textbf{ Gemini-1 2 GPUs (56 sec)}}\\ \hline
        Tot.  Energy reported &  1.689 & 1.630 & 4.570 & 1.640 & 1.620 & 2.350 & N/A & ~ & ~ & ~ \\ \hline
        Tot.  Energy w/o PUE & 1.008 & 1.630 & 4.570 & 1.640 & 1.620 & 1.516 & N/A & 9.333 & 3.729 & N/A \\ \hline
        Energy for  CPU &  0.130 & 0.000 & 0.000 & - & - & - & ~ & - & - & ~ \\ \hline
        Energy for  GPU & 0.139 & 1.620 & 1.620 & - & - & - & ~ & - & - & ~ \\ \hline
        Energy for  Memory & 0.739 & 0.013 & 2.950 & - & - & - & ~ & - & - & ~ \\ \hline
        Carbon emissions & 0.086 & 0.090 & 0.250 & 0.110 & 0.110 & 0.140 & ~ & 0.622 & 0.440 \\ \hline

        \multicolumn{11}{l}{  \textbf{ Rosenblatt (1min 36 sec)}}\\ \hline
        Tot.  Energy reported & 1.030 & 3.190 & 3.800 & 2.000 & 2.100 & 4.56 & 3.860 & 13.30 & 6.711 &  \\ \hline
        Tot.  Energy w/o PUE & 0.617 & 3.190 & 3.800 & 2.000 & 2.100 & 2.940 & 2.440 & 13.30 & 6.711 & N/A \\ \hline
        Energy for  CPU & 0.148 & 1.200 & 1.200 & - & - & - & - & - & - &  \\ \hline
        Energy for  GPU & 0.086 & 1.900 & 1.900 & - & - & - & - & - & - &  \\ \hline
        Energy for  Memory & 0.389   & 0.0276 & 0.600 & - & - & - & - & - & - &  \\ \hline
        Carbon emissions & 0.0527 & 0.170 & 0.200 & 0.138 & 0.140 & 0.266 & 0.210 & 0.533 & 0.792 &  \\ \hline

        \multicolumn{11}{l}{  \textbf{ Labri (45 sec)}}\\ \hline
        Tot. Energy reported & 1.94 & 1.689 & 2.287 & 1.1459 & 1.126 & 2.219 & 1.91 & 9.375 & 2.093 & ~  \\ \hline
        Tot. Energy w/o PUE & 1.16 & 1.689 & 2.287 & 1.1459 & 1.126 & 1.432 & 1.209 & 9.375 & 2.093 & 2.241 \\ \hline
        Energy for CPU & 0.255 & 0.565 & 0.565 & - & - & - & -  & - & - & - \\ \hline
        Energy for GPU & 0.128 & 1.111 & 1.097 & - & - & - & -  & - & - &  - \\ \hline
        Energy for Memory & 0.777 & 0.013 & 0.626 & - & - & - & -  & - & - & - \\ \hline
        Carbon emissions & 0.099 & 0.093 & 0.126 & 0.074 & 0.076 & 0.13 & 0.107 & 0.375 & 0.247 &  \\ \hline

        \multicolumn{11}{l}{  \textbf{ Personal computer (57 sec) }}\\ \hline
        Tot.  Energy reported & 0.356 & 1.000 & 1.190 & 0.733 & 0.728 & 1.415 & N/A & 4.167 & 3.949 &  \\ \hline
        Tot.  Energy w/o PUE & 0.356 & 1.000 & 1.190 & 0.733 & 0.728 & 0.913 & N/A & 4.167 & 3.949 & 1.404 \\ \hline
        Energy for  CPU & 0.032 & 0.330 & 0.330 & - & - & - &  & - & - & - \\ \hline
        Energy for  GPU & 0.125 & 0.660 & 0.660 & - & - & - &  & - & - & -\\ \hline
        Energy for  Memory & 0.199 & 0.015 & 0.195 & - & - & - &  & - & - &  -\\ \hline
        Carbon emissions & 0.018 & 0.056 & 0.065 & 0.049 & 0.049 & 0.083 &  & 0.167 & 0.466  & \\ \hline

        \multicolumn{11}{l}{  \textbf{ Colab - Oregon (1 min 6 sec)}}\\ \hline
        Tot.  Energy reported & 0.381 & 1.500 & 1.600 & 3.000 & 3.000 & 0.805 & N/A & 1.280 & 5.15 &  \\ \hline
        Tot.  Energy w/o PUE & 0.343 & 1.500 & 1.600 & 3.000 & 3.000 & 0.519 & N/A & 1.280 & 5.15 &   N/A \\ \hline
        Energy for  CPU & 0.219 & 0.900 & 0.900 & - & - & - &  & - & - &  \\ \hline
        Energy for  GPU & 0.041 & 0.600 & 0.600 & - & - & - &  & - & - &  \\ \hline
        Energy for  Memory & 0.0913 & 0.0206 & 0.100 & - & - & - &  & - & - &  \\ \hline
        Carbon emissions& 0.024  & 0.200 & 0.200 & 0.600 & 0.600 & 0.290 &  & 0.367 & 1.03 &\\ \hline
    \end{tabular}%
    }
    \caption{Results for the training of a digit classifier (experiment 1). All consumption values are in Wh. Carbon emissions are in gCO2e. 
    For CodeCarbon and Eco2AI, (P) refers to the process tracking mode and (M) to the machine tracking mode.}
    \label{ResultsExp1}
\end{table}

\begin{table}[!h]
    \centering
    \resizebox{\textwidth}{!}{%
    
    \begin{tabular}{ L{4.2cm}|C{2.2cm} C{2.2cm} C{2.2cm} C{2.2cm} C{2.2cm} C{2.2cm} C{2.2cm} C{2.2cm} C{2.2cm}C{2.2cm}}
    \hline
    
        ~ & Green-Algorithms & CodeCarbon (P) & CodeCarbon (M) & Eco2AI (P) & Eco2AI (M) & CarbonTracker & EIT & MLCO2 & Cumulator & Wattmeter \\ \hline

        \multicolumn{11}{l}{  \textbf{ Gemini-1 whole node (2 hs)}}\\ \hline
        Total Energy reported & 1.92 & 1.39 & 1.69 & 1.07 & 1.10 & 2.09 & 2.09 & 4.80 & 0.5 &  \\ \hline
        Tot. Energy w/o PUE & 1.15 & 1.39 & 1.69 & 1.07 & 1.10 & 1.35 & 1.32 & 4.80 & 0.5 & 2.10 \\ \hline
        Energy for CPU & 0.09 & 0.22 & 0.22 & - & - & - & - & - & - &  -\\ \hline
        Energy for GPU & 0.69 & 1.14 & 1.09 & - & - & - & - & - & - &  -\\ \hline
        Energy for Memory & 0.37 & 0.03 & 0.37 & - & - & - & - & - & -  &  -\\ \hline
        Carbon emissions  & 100 & 80 & 90 & 70 & 80 & 120 & 120 & 280 & 60 &  \\ \hline
        
        \multicolumn{11}{l}{  \textbf{ Gemini-1 2 GPUs (1h 17 min)}}\\ \hline
        Total Energy reported & 0.76 & 0.36 & 0.61 & 0.35 & 0.37 & 0.59 & N/A & 0.77 & 0.45 &  \\ \hline
        Tot. Energy w/o PUE & 0.47 & 0.36 & 0.61 & 0.35 & 0.37 & 0.38 & N/A & 0.77 & 0.45 & N/A  \\ \hline
        Energy for CPU &  0.05 & 0 & 0.00 & - & - & - &  & - & - &  \\ \hline
        Energy for GPU & 0.36 & 0.359 & 0.37 & - & - & - &  & - & - &   \\ \hline
        Energy for Memory & 0.06 & 0.008 & 0.24 & - & - & - &  & - & - &  \\ \hline
        Carbon emissions & 40 & 20 & 34 & 24 & 25 & 34 &  & 51 & 38 &  \\ \hline
        
        \multicolumn{11}{l}{  \textbf{  Rosenblatt (3hs 16 min) }}\\ \hline
        Tot. Energy reported & 1.77 & 1.07 & 1.12 & 0.89 & 0.99 & 1.71 & 1.75 & 1.63 & 0.84 &  \\ \hline
        Tot. Energy w/o PUE & 1.06 & 1.07 & 1.12 & 0.89 & 0.99 & 1.10 & 1.11 & 1.63 & 0.84 & N/A \\ \hline
        Energy for CPU & 0.10 & 0.17 & 0.17 & - & - & - & - & - & - &  \\ \hline
        Energy for GPU & 0.88  & 0.89 & 0.87 & - & - & - & - & - & - &  \\ \hline
        Energy for Memory & 0.08 & 0.02 & 0.08 & - & - & - & - & - & - &  \\ \hline
        Carbon emissions & 90 & 60 & 60 & 60 & 70 & 100 & 100 & 90 & 100 &  \\ \hline
        
        \multicolumn{11}{l}{  \textbf{  Labri (1h 13 min)}}\\ \hline
        Tot.  Energy reported & 0.80 & 0.76 & 0.79 & 0.69 & 0.72 & 1.16 & 1.17 & 0.9 & 0.3 \\ \hline
        Tot.  Energy w/o PUE & 0.48 & 0.76 & 0.79 & 0.69 & 0.72 & 0.75 & 0.74 & 0.9 & 0.3 & 0.83 \\ \hline
        Energy for  CPU & 0.15 & 0.097 & 0.097 & - & - & - & - & - & -  & -\\ \hline
        Energy for  GPU & 0.27 & 0.66 & 0.64 & - & - & - & - & - & -  & -  \\ \hline
        Energy for  Memory & 0.06 & 0.03 & 0.056 & - & - & - & - & - & - & - \\ \hline
        Carbon emissions & 41 & 42 & 44 & 47 & 48 & 68 & 65 & 36 & 24 \\ \hline
        
        \multicolumn{11}{l}{  \textbf{ Personal computer (1h 49 min)}}\\ \hline
        Tot. Energy reported & 0.37 & 0.34 & 0.35 & 0.25 & 0.27 & 0.52 & N/A & 0.45 & 0.46 &  \\ \hline
        Tot. Energy w/o PUE & 0.37 & 0.34 & 0.35 & 0.25 & 0.27 & 0.34 & N/A & 0.45 & 0.46 & 0.40 \\ \hline
        Energy for CPU &  0.001 & 0.09 & 0.09 & - & - & - &  & - & - & - \\ \hline
        Energy for  GPU & 0.35 & 0.24 & 0.24 & - & - & - &  & - & - &  -\\ \hline
        Energy for  Memory & 0.02 & 0.01 & 0.02 & - & - & - &  & - & - & - \\ \hline
        Carbon emissions  & 19 & 19 & 19 & 17 & 18 & 30 &  & 18 & 54 &  \\ \hline

        \multicolumn{11}{l}{  \textbf{ Colab - Oregon (17 hs est.) }}\\ \hline
        Tot. Energy reported & 1.22 & 1.49 & 1.56 & 1.03 & 1.82 & 0.96  & N/A & 1.19 & 0.36 &  \\ \hline
        Tot. Energy w/o PUE & 1.10 & 1.49 & 1.56 & 1.03 & 1.82 & 0.62 & N/A & 1.19 & 0.36 &  N/A\\ \hline
        Energy for  CPU & 0.07 & 0.73 & 0.73 & - & - &  	&  ~ & - &  \\ \hline
        Energy for  GPU & 0.95 & 0.75 & 0.75 & - & - & - &  & - & - &  \\ \hline
        Energy for  Memory & 0.08 & 0.02 & 0.08 & - & - & - &  & - &  &  \\ \hline
        Carbon emissions & 199 & 206 & 216 & 184 & 328 & 369 &  & 100 & 72	& \\ \hline
        
    \end{tabular}%
    }
    \caption{Results for the training of an image denoiser (experiment 2). All consumption values are in kWh. Carbon emissions are in gCO2e. The consumption indicated for Colab is extrapolated. An epoch was executed, the consumptions were obtained, and the values were extrapolated.}
    \label{ResultsExp2}
\end{table}

\subsubsection{Variability between the different tools\\}\label{variability}

From the two tables~\ref{ResultsExp1} and~\ref{ResultsExp2}, we observe a large difference between the energy consumption and carbon emissions reported by the different tools. For instance, a 400\% increase of consumption for MLCO2 compare to Eco2AI on the Gemini-1 node of Grid5000.

\paragraph{Machine vs Process} 

Some tools are focused on estimating the consumption of the entire machine, and are comparable with wattmeters, but others estimate the consumption of the process, trying to isolate it from other processes that may be running on the machine.

CodeCarbon and CarbonTracker have similar strategies for GPU and CPU consumption estimation, focusing on full machine estimation. They differ in method for the estimation of memory consumption. We can say that CodeCarbon strategy is more accurate, since it reaches a value more similar to that of the wattmeter.

Eco2AI and EIT focus more on isolating the consumption of the process that is measured. It can be seen from both experiments that these tools show a lower consumption estimate than CodeCarbon and CarbonTracker. Green-Algorithms approach also attempts to isolate consumption from the process.

\paragraph{Multiple GPUs}  Cumulator only measures CPUs or GPUs, according to what we specify when creating the tracker. In both cases it considers a single unit of the hardware it is measuring, without checking how many CPUs or GPUs exist on the machine. 
 
MLCO2 also has a simplified view, only measuring the consumption of 1 GPU. The values reported in the tables were obtained by multiplying the value obtained by the number of GPUs available. The reported values for 1 GPU for Cumulator and MLCO2 are very similar because they follow the same strategy. In the case of the personal computer or Colab, having a single GPU, we can come to consider these two tools, but we are also not measuring CPU consumption. In addition, the tools only multiply the time consumed by the TDP, so it does not verify actual consumption or compute usage factors. Their results can only be useful when we have a single unit of the component to be measured (CPU or GPU for Cumulator, and only GPU for MLCO2), and it has a usage factor close to 100\%.

\paragraph{Usage factor} 
The web calculator of Green-Algorithms and their server tool G4HPC set default usage factor to 100\% CPU and GPU loads if these data are not provided. This will overestimate power consumption in most cases. To be considered by GA, usage factors must be calculated by the user. The CPU usage factor can be calculated using the CPU time and the process time, but there is no easy way to get the GPU usage factor. We can get empirical values from measurements using \texttt{nvidia-smi} while the algorithm is running, and assume that it maintains that usage factor throughout the run. In this case we are assigning all the utilization percentage reported by \texttt{nvidia-smi} to the process, but there could be other processes using the GPU. In our study, since for both experiments the only process running on GPUs was the one measured, we took one sample per epoch of the \texttt{nvidia-smi} output during code execution. We averaged the utilization percentage values of all GPUs across all samples. Results are shown in table \ref{usageFactor}. We observe a low usage factor, especially on servers.  As shown in table \ref{ResultsExp2}, MLCO2 seems to largely overestimate consumption on Grid5000, which is because it does not take into account the usage factor of the GPUs. The average usage factor is 14\%, but MLCO2 is considering 100\% for all 8 GPUs. 
EIT queries and calculates usage factors during execution. Eco2AI only does this for CPU, as it directly queries the consumed energy for GPU. CodeCarbon and CarbonTracker directly query the consumed energy for both GPU and CPU, without using the usage factor.

\begin{table}[!h]
    \caption{Usage factor of CPU and GPU in the infrastructures used. This computed values are used by Green-Algorithms.\label{usageFactor}}
    \centering
    
    \begin{tabular}{l|cccc}
    \hline
        ~ & CPU & GPU & CPU & GPU \\ 
        ~ & Expe. 1 & Expe. 1 & Expe. 2 & Expe. 2 \\ \hline
        Gemini-1 (Grid5000) & 5\% & 0.3\% & 16\% & 14\% \\ \hline
        Gemini-1  2 GPUs (Grid5000) & 12\% & 1.5\% & 58\% & 46\% \\ \hline
        Server (Labri) & 9\% & 1\% & 73\% & 35\% \\ \hline
        Rosenblatt (MAP5) & 16\% & 1\% & 39\% & 54\% \\ \hline
        Personal Computer & 22\% & 3\% & 4\%  & 77\% \\ \hline
    \end{tabular}
    
\end{table}


\subsubsection{Comparison between software tools and wattmeter\\}\label{res-watt}

Wattmeters were present in Labri server, the personal computer and Gemini-1. Table \ref{ComparisionExperiment} shows a summary of the comparison presented. For experiment 1, wattmeter on the personal computer and labri server only made one measurement during the entire experiment, so the reported value may not be exact.

In the first experiment, the value reported by the consumption of the machine with CodeCarbon is almost exactly the same as that reported by the wattmeter. For the second experiment the value is not as precise, but it is still more than 80\% for all infrastructures. This measuring tool is the one that gives the closest value with respect to wattmeters, followed by CarbonTracker, with more variability between infrastructures.

Eco2AI and EIT report values larger than the wattmeter. Since these tools try to isolate the consumption of the process, and not measure the total consumption of the machine, then the reports of energy consumption are not comparable with the wattmeter value.

\begin{table}[!h]
    \centering
    \setlength{\tabcolsep}{0.1cm} 
    
    \begin{tabular}{ L{4.4cm}|C{2.75cm} C{2.7cm} C{2.3cm} C{2.7cm}}
    \hline
        ~ & CodeCarbon~(M) & Eco2AI (M) & CarbonTracker  & EIT\\ \hline
        Expe. 1 Grid5000 & 96\% & 55\% & 66\% & 13\% \\ \hline
        Expe. 2 Grid5000 & 80\% & 60\% & 64\% & 63\% \\ \hline
        Expe. 1 Personal comp. & 85\% & 52\% & 65\% & N/A \\ \hline
        Expe. 2 Personal comp. & 88\% & 68\% & 85\% & N/A \\ \hline
        Expe. 1 Labri & 102\% & 50\% & 64\% & 54\% \\ \hline
        Expe. 2 Labri & 95\% & 87\% & 90\% & 89\% \\ \hline
    \end{tabular}
    \caption{Comparison between software tools and wattmeter in Grid5000 (without considering PUE), personal computer and Labri server. Values represent the percentage of energy reported by tools wrt the value reported by the wattmeter.}
    \label{ComparisionExperiment}
\end{table}

\subsection{Influence of infrastructures}\label{res-infra}

We ran the same experiments on different infrastructures. For both experiments, power consumption is higher on larger infrastructures (e.g. Gemini-1).

As an example, the Denoiser training experiment took 2 hours on Gemini-1 (Grid5000 server), while on Rosenblatt (MAP5 server) it took 3 hours and 16 minutes. Usage factor of CPU was lower in Grid5000: 16\% in Grid5000 and 39\% in MAP5. The estimation of usage factor of GPU was also lower in Grid5000: 14.3\% in Grid5000 while in MAP5 it was 54\%.
The consumption reported in Gemini-1 by CodeCarbon (Machine tracker) is 1.69 kWh, while the consumption reported in Rosenblatt by CodeCarbon (Machine tracker) was 1.12 kWh. Rosenblatt's hardware is considerably smaller than Gemini-1's (see table \ref{InfraSpecs}). 

It can also be seen that in experiment 2 for Labri, the personal computer and on Gemini-1 booking only 2 GPUs, the execution time was less than in the case of execution on the entire Gemini-1 node. This longer execution is more likely due to the parallelization strategy (using nn.DataParallel) that runs 
the training on all GPUs without requiring their full computing power. 

This might be a good reason for using, as much as possible, a hardware which size is adapted to the experiments where resources can be used as much as possible, even if the experiments take more time. Gemini-1 node has 8 GPUs which is not useful for both our experiments.

\subsection{Data load}\label{res-data}

In the Denoiser training experiment, we seperately quantified the energy consumption of data loading (6GB Imagenet validation split) vs training the model and found that only 0.5\% of the energy was used to load the data. This is partly because the data was already on the server, the impact of downloading the data and of data storage is not being measured.

\subsection{Batch size}\label{res-batch}

To study the impact of batch size during training, we used CodeCarbon during experiment 2 (denoiser) on the Gemini-1 node for 10 epochs. Using three batch sizes (32, 64 and 128), we showed that there is a tradeoff between energy used and runtime (Table \ref{ExpBAtch}). While larger batch sizes led to faster runtimes, the largest energy usage was measured for the smallest batch size (32), closely followed by the largest one (128). In this situation, an intermediate batch size of 64 looks like a better compromise, combining a runtime not far off the shortest one and minimising energy usage.

However, when we decrease the batch more, the experiment takes longer, and the idle consumption of the resources starts to weigh on the total consumption of the experiment. If we compare the GPU consumption of experiments with batch size 32 and 128, we see that experiment 32 consumes less, still taking almost 3 times longer. Nevertheless, comparing the experiment of 32 with that of 64, we have that the consumption is higher, probably because the experiment takes almost 10 minutes more, and we have the static consumption of the resources.

In conclusion, a balance is required between the length of the experiment, and the greater consumption of the GPU memory to obtain a minimum energy consumption.

\begin{table}[!h]
    \centering
    \setlength{\tabcolsep}{0.1cm} 
    \begin{tabular}{ L{5.1cm}|C{3.4cm} C{3.4cm} C{3.4cm} }
    \hline
        ~ & Experiment with batch size 32 & Experiment with batch size 64 & Experiment with batch size 128 \\ \hline
        Total Energy (CodeCarbon)& 252 & 184 & 246 \\ \hline
        CPU (CodeCarbon)& 41 & 29 & 20 \\ \hline
        GPU (CodeCarbon)& 205 & 152 & 224 \\ \hline
        Memory (CodeCarbon)& 6 & 3 & 2.3 \\ \hline
        Total Energy (Wattmeter)& 391 & 280.3 & 320 \\ \hline
        Time spent & 25:54 & 16:29 & 10:30 \\ \hline
    \end{tabular}
    \caption{Results of experiment 2 with different batch sizes. All consumption values are in Wh.}
    \label{ExpBAtch}
\end{table}

\subsection{Checkpoints}\label{res-check}

We found that checkpointing had no impact on energy consumption or runtime (Table \ref{ExpCheckpoint}). We tested this on experiment 2 on Gemini-1 using CodeCarbon and a wattmeter. In the first scenario, the values of the network parameters were saved every epoch (ten epochs in total) and in the second scenario values were saved only once.

\begin{table}[!h]
    \centering
    \setlength{\tabcolsep}{0.1cm} 
    \begin{tabular}{ L{7cm}|C{3.5cm} C{3.5cm} }
    \hline
        ~ & Experiment with one checkpoint & Experiment with ten checkpoints \\ \hline
        Total Energy reported (CodeCarbon) & 161 & 160 \\ \hline
        Energy for CPU (CodeCarbon) & 24 & 24 \\ \hline
        Energy for GPU (CodeCarbon) & 134 & 133 \\ \hline
        Energy for Memory (CodeCarbon) & 3 & 3 \\ \hline
        Total Energy reported (Wattmeter) & 206 & 206 \\ \hline
        Time spent (min) & 14:10 & 13:47 \\ \hline
    \end{tabular}
    \caption{Results of experiment with different frequency of checkpoints. Both experiments are run for 10 epochs. On the left column, only one checkpoint has been saved at the end of these epochs. On the right column, one checkpoint is saved per epoch. All consumption values are in Wh.}
    \label{ExpCheckpoint}
\end{table}

\subsection{Variability of consumption through epochs} 

\begin{figure}[htbp]
\centering
\includegraphics[width=0.7\textwidth]{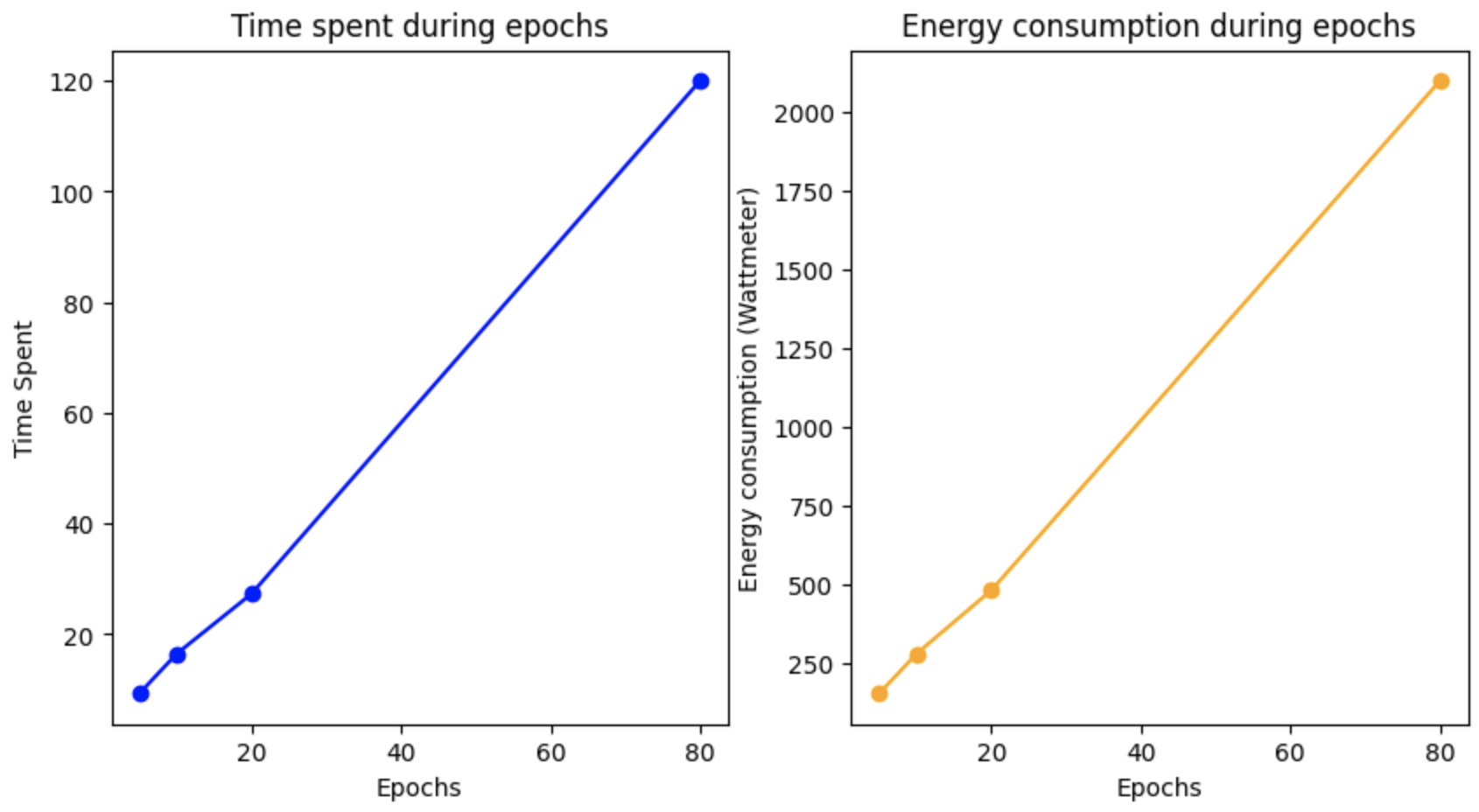}
\caption{Duration and energy consumption after different number of epochs of Experiment 2. All consumption values are in Wh.}
\label{GraphicsExtrapolation}
\end{figure}

It is interesting to determine if it is possible to extrapolate the energy consumption of a training phase from the values observed on only few epochs.
To determine it, the Denoiser training experiment was executed during different number of epochs on Gemini-1; the time consumed was measured, as well as the energy consumption.  
Results in the Fig. \ref{GraphicsExtrapolation} show that epochs duration and consumption are constant. It might therefore be possible to extrapolate energy consumption for large experiments from experiments on just a few epochs. Same conclusion was reached in \cite{CarbonTracker} when using CarbonTracker.

\subsection{Is measuring really eco-friendly?} \label{res-tools}

To compare the extra energy consumption of the tools themselves, we run 2 processes of experiment 2 in parallel, one with all seven trackers and one without any. We report energy consumption provided by the wattmeter.   
We found that the code with trackers was almost 10\% slower and ended 11 minutes later than the one without trackers. The energy consumption during this extra time was  0.19 kWh, while it was 2.58 kWh for the time when both processes were running in parallel (+7.4\%).

Another experiment was performed, testing each tracker at a time. 
As in previous test, we run two processes in parallel, one with a given tracker and one without any tracker. Energy is measured with wattmeter.

Table \ref{2ProcessExp2} shows the result with 10 epochs. It can be seen that the additional energy is around 1\% of the total consumption for all the tools, except for Eco2AI, where consumption reaches 3.5\%, a value that is not negligible. We think that the biggest consumption compared to the other tools is not using the RAPL files to obtain the memory and CPU consumption, but rather making queries to the operating system to later do the calculations. Although other tools also do it this way, none do it to calculate the energy of both resources.

\begin{table}[!h]
    \centering
    \resizebox{\textwidth}{!}{%
    \setlength{\tabcolsep}{0.0cm} 
    \begin{tabular}{ L{5.5cm}| C{2.8cm} C{2.3cm} C{2.3cm} C{1.9cm} C{2.3cm}}
    \hline
        ~ & CodeCarbon(P)& Eco2AI(P) & CarbonTracker & EIT & Cumulator \\ \hline
        Run time  w/ tracker (min) & 15:09 & 15:33 & 16:35 & 16:29 & 15:02 \\ \hline
        Run time  w/o tracker  (min) & 15:05 & 14:57 & 16:24 & 16:35 & 14:49 \\ \hline
        Extra time  with tracker  (min)& 0:04 & 0:36 & 0:11 & -0:06:00 & 0:13 \\ \hline
        Energy Cons.when 2 processes running (Wh) & 335.5 & 334 & 358 & 358.5 & 331.6 \\ \hline
        Energy Cons. during extra time (Wh) & 3.1 & 12.2 & 4.29 & 0 & 5.4 \\ \hline
        Percentage of overload (\%)& 0.92 & 3.5 & 1.2 & 0 & 1.6 \\ \hline
    \end{tabular}%
    }
    \caption{Results running experiment 2 twice in parallel on Gemini-1: one process using trackers, the other without.}
    \label{2ProcessExp2}
\end{table}%

As a result of both experiments, we can conclude that measuring the processes has an impact, but a small one. The first experiment carried out with all the trackers has a longer execution time, probably due to delays while access to resources. It might be a good idea to use online tools such as Green-Algorithms, in order not to add additional load to the algorithm and still being able to measure the impact. 

\subsection{Static and deployment consumption} \label{res-idle}
All the tools discussed in this guide are limited to quantifying energy consumption while training a deep learning approach. But infrastructures also us energy when nodes are not used or when the final solution is deployed. The authors of \cite{luccioni2022estimating} studied static infrastructure emission and deployment emissions when training BLOOM, a large language model and found these to be substantial.

We measured the energy consumption of idle resources on Gemini-1 over the same period of time it takes to run experiment 1. In an idle situation, no process is being run  beyond those required by the operating system. We performed the same procedure with experiment 2 (executed for 10 epochs). The results are shown in the table \ref{IdleExp}. Idle energy consumption is around 745Wh. We see that the consumption of idle resources is high comparing with the consumption reported during training: 84.4\% for experiment 1 and 72.9\% for experiment 2. Note that for both experiments, the resources are not fully used. In the table \ref{usageFactor} we can see the percentage of CPU and GPU utilization during.

\begin{table}[!ht]

    \caption{Static (Idle) and dynamic energy consumption measured with wattmeters\label{IdleExp}}
    \centering
    \begin{tabular}{l|c|c}
    \hline
        ~ & Time (min) & Energy consumption (Wh) \\ \hline
        Experiment 1 & 00:53 & 12.96  \\ \hline
        Idle         & 00:53 & 10.95  \\ \hline
        Experiment 2 (10 epochs) & 16:29 & 280.3 \\ \hline
        Idle                     & 16:29 & 204.4  \\ \hline
    \end{tabular}
\end{table}

This result is interesting since we can see that most of the consumption occurs simply by having the hardware available to use it. This tells us that we have to be very careful when leaving hardware on for availability. The availability and immediacy of resources is very expensive in terms of energy consumption.
When we use hardware where we do not have the power to turn it off when we are not using it, such as the cloud, or shared computers, we must remember that there is an additional consumption to be able to make a reservation at any time for a given resource.

\section{Discussions}\label{discussions}
This section summarizes our observations and anticipates questions that AI practitioners may have when starting to measure the energy consumption of their codes. 

\subsection{When to measure impacts?}

Contrary to tracking tools, online ones like Green-Algorithms make it possible to estimate consumption both after training, as concluded in \cite{bannour2021evaluating}, and before training. Although this will be less precise, it anticipating the environmental impacts of a project.

If we use software tools and perform more than one run, we recommend performing the measurement only for some runs.

Given that is possible to extrapolate the energy consumption of a training phase from the values observed on only few epochs, we could measure the consumption of the firsts epochs, and then estimate the consumption of the total training. In this way, the consumption corresponding to the measurement will be slightly lower.

\subsection{Which tools to use}

Estimating power consumption using software tools adds small load, so it might be a good idea to use online tools like Green-Algorithms. 

Green-Algorithms is the most versatile tool, as it can be used under different infrastructures, brands of CPUs and GPUs. However, online tools requires manual intervention to obtain the information and may be less precise. A first step to remedy this is the tool \href{https://www.green-algorithms.org/GA4HPC/}{GA4HPC} which is used to obtain the resource reservation data of a job in clusters that use SLURM as workload manager.

MLCO2 is an online tool but is much more limited. It just account for GPU consumption and the value returned must be correctly weighted according to the number of GPUs and the correct execution time of the algorithm.

If we want to use software tools, we found that CodeCarbon is the best tool among those studied to estimate the total consumption of the machine. The consumption reported with it is more accurate when accessing RAPL files. However, a strength of this tool is that it can be used without access to them. 
On the contrary, if what you want is to isolate the consumption of the process using software tools, Eco2AI and EIT are those that try to do it. Eco2AI does not require access to the RAPL files and is maintained and updated. By contrast, EIT requires access to the RAPL files and it is necessary to modify the code to use the tool.

\subsection{Which infrastructure to use}

Since the idle consumption of resources is a large percentage of the total consumption, we recommend only keeping available the resources needed to achieve a high usage factor and have the minimum idle consumption, even if the execution time is longer. 

With supercomputers, we recommend requesting only the necessary resources, and if it is adequate, to share the infrastructure with other user processes.

If possible, we recommend turning off personal computers or servers as soon as computation is done.

If we are using cloud infrastructure, as far as possible, choose data centers that have the lowest PUE and that are located in areas with low gas emissions. We recommend choosing low emission hours for the execution of training. Carbon aware schedulers such as \href{https://github.com/GreenScheduler/cats}{CATS}, \href{https://github.com/thegreenwebfoundation/grid-intensity-go}{grid-intensity-go} or \href{https://pypi.org/project/carbon-aware-scheduler/}{carbon-aware-scheduler} can be used to help with this.

\subsection{Other impacts}\label{sec:data}
In this paper, we have been focusing only on energy consumption, and associated greenhouse gas emissions, for training AI models. This only a small part of total energy consumption of the complete life cycle of the AI service. 

For the training phase, an AI practitioner generally trains the model several times. Complete training emissions should consider all runs. In Green-Algorithms, we can model multiple runs, associated with retraining using the "pragmatic scaling factor" parameter. 

As mentioned in previous studies \cite{bannour2021evaluating, luccioni2022estimating,wu_sustainable_2021},  the energy consumption is underestimated, since all the tools only measure the consumption during training and not during deployment. Studies \cite{wu_sustainable_2021,luccioni2022estimating} have measured the consumption of deployment phases that can be much higher than the one of training. Here again, choosing appropriate resources to have a high usage factor seems to be essential. 

Many other environmental impacts (resource depletion, ecotoxicity, etc.) linked to the life cycle of equipments (manufacturing, transport, distribution, use, end of life), are here not discussed and should be investigated. Even of carbon footprint, computing embodied emissions is a challenge since all data are not made public by manufacturers. From several assumptions, the authors of~\cite{luccioni2022estimating} propose an estimation of embodied emissions equal to half the ones of training. \\

Datasets creation, transfer and storage  are also very important aspects of AI. An estimate by \cite{malmodin2016energy} is 0.023 kWh/GB for transferring data on the IP core. For storage, there are various estimates. Following Seagate measurement \footnote{\href{https://www.seagate.com/gb/en/global-citizenship/product-sustainability/}{https://www.seagate.com/gb/en/global-citizenship/product-sustainability/}}, \cite{lannelongue2023carbon} consider an order of magnitude of the carbon footprint of storing 1 terabyte of data to be around 10 kgCO2e per year. Another study \cite{groger2021green} mention 52 Wh for storing one gigabyte for one year. 
To know more about energy management techniques for database systems, we refer the reader to the systematic review~\cite{guo2022energy}.

\subsection{Predicting impacts} 
Systematically estimating the carbon footprint of AI project can raise awareness, encourage the development of energy-efficient software and  limit the waste of resource~\cite{lannelongue2023carbon}. Importantly, these impacts should be anticipated before the start of a project. Authors of \cite{lefevre2023environmental} propose a list of criteria for assessing the environmental impacts of projects involving Artificial Intelligence (AI) methods. In addition to measuring while training or deploying an AI model, AI users should try to anticipate as much as possible the impacts of their computations are likely to have, as well as the behavioral, economic, or societal changes that might be induced by the project. In the same line, \cite{WILSON2022101926} review ethics, explainability, responsibility, and accountability concepts in AI and propose a model for sustainable AI in the public sector.

\section{Conclusion}
In this paper we have presented and analyzed seven existing tools for estimating energy consumption when training a deep learning model. We have explained the specificities of each tool and detailed the notions that may be not well known by AI practitioners. From our study, we have drawn some analysis and recommendations in previous sections. 
Remark that our two experiments were related to training regular CNNs for image processing and analysis. We believe that the main results would hold for other types of architectures, as carbon footprint estimators have shown the same behaviors for other applications or workloads in \cite{jay:hal-04030223, bannour2021evaluating, strubell22}. 
In the paper we have highlighted the advantages and limits of online tools, and that the choice of the software tool depends on the infrastructure and on either one wants to measure the whole node or the process only. We have also shown that measuring with software tools has a small impact that can become non negligible for large experiments. We observed that consumption is constant through epochs, and therefore measuring only on few epochs and extrapolating can be sufficient. We have confirmed that it is important to train models on infrastructures that is scaled to the need, not booking a whole node when not necessary. Finally, all these tools measure only dynamic energy consumption of computing and further studies are required to include static consumption and environmental impacts.

\section{Acknowledgments}
This study has been carried out with financial support from the French Research Agency through the PostProdLEAP project (ANR-19-CE23-0027-01). 
Loïc Lannelongue was supported by core funding from the British Heart Foundation (RG/18/13/33946); the NIHR Cambridge Biomedical Research Centre (BRC-1215-20014; NIHR203312)[*]; the Cambridge British Heart Foundation Centre of Research Excellence (RE/18/1/34212); and the BHF Chair Award (CH/12/2/29428). *The views expressed are those of the authors and not necessarily those of the NIHR or the Department of Health and Social Care.
The authors thank Michael Clément and Boris Mansencal for running experiments in Labri and personal computer. We also thank Mathilde Jay, Denis Trystram, Laurent Lefèvre and Anne-Laure Ligozat for fruitful discussions.


\bibliographystyle{IEEEtran}
\bibliography{bibliography.bib}

\begin{thebibliography}{10}
\providecommand{\url}[1]{#1}
\csname url@samestyle\endcsname
\providecommand{\newblock}{\relax}
\providecommand{\bibinfo}[2]{#2}
\providecommand{\BIBentrySTDinterwordspacing}{\spaceskip=0pt\relax}
\providecommand{\BIBentryALTinterwordstretchfactor}{4}
\providecommand{\BIBentryALTinterwordspacing}{\spaceskip=\fontdimen2\font plus
\BIBentryALTinterwordstretchfactor\fontdimen3\font minus
  \fontdimen4\font\relax}
\providecommand{\BIBforeignlanguage}[2]{{%
\expandafter\ifx\csname l@#1\endcsname\relax
\typeout{** WARNING: IEEEtran.bst: No hyphenation pattern has been}%
\typeout{** loaded for the language `#1'. Using the pattern for}%
\typeout{** the default language instead.}%
\else
\language=\csname l@#1\endcsname
\fi
#2}}
\providecommand{\BIBdecl}{\relax}
\BIBdecl

\bibitem{rolnick_tackling_2019}
\BIBentryALTinterwordspacing
D.~Rolnick, P.~L. Donti, L.~H. Kaack, K.~Kochanski, A.~Lacoste, K.~Sankaran,
  A.~S. Ross, N.~Milojevic-Dupont, N.~Jaques, A.~Waldman-Brown, S.~Luccioni,
  T.~Maharaj, E.~D. Sherwin, S.~K. Mukkavilli, K.~P. Kording, C.~Gomes, A.~Y.
  Ng, D.~Hassabis, J.~C. Platt, F.~Creutzig, J.~Chayes, and Y.~Bengio,
  ``\BIBforeignlanguage{en}{Tackling {Climate} {Change} with {Machine}
  {Learning}},'' \emph{\BIBforeignlanguage{en}{arXiv:1906.05433 [cs, stat]}},
  Nov. 2019, arXiv: 1906.05433. [Online]. Available:
  \url{http://arxiv.org/abs/1906.05433}
\BIBentrySTDinterwordspacing

\bibitem{vinuesa_role_2020}
\BIBentryALTinterwordspacing
R.~Vinuesa, H.~Azizpour, I.~Leite, M.~Balaam, V.~Dignum, S.~Domisch,
  A.~Felländer, S.~D. Langhans, M.~Tegmark, and F.~Fuso~Nerini,
  ``\BIBforeignlanguage{en}{The role of artificial intelligence in achieving
  the {Sustainable} {Development} {Goals}},''
  \emph{\BIBforeignlanguage{en}{Nature Communications}}, vol.~11, no.~1, p.
  233, Dec. 2020. [Online]. Available:
  \url{http://www.nature.com/articles/s41467-019-14108-y}
\BIBentrySTDinterwordspacing

\bibitem{kar_how_2022}
\BIBentryALTinterwordspacing
A.~K. Kar, S.~K. Choudhary, and V.~K. Singh, ``\BIBforeignlanguage{en}{How can
  artificial intelligence impact sustainability: {A} systematic literature
  review},'' \emph{\BIBforeignlanguage{en}{Journal of Cleaner Production}},
  vol. 376, p. 134120, Nov. 2022. [Online]. Available:
  \url{https://linkinghub.elsevier.com/retrieve/pii/S0959652622036927}
\BIBentrySTDinterwordspacing

\bibitem{strubell_energy_2019}
\BIBentryALTinterwordspacing
E.~Strubell, A.~Ganesh, and A.~McCallum, ``\BIBforeignlanguage{en}{Energy and
  {Policy} {Considerations} for {Deep} {Learning} in {NLP}},''
  \emph{\BIBforeignlanguage{en}{arXiv:1906.02243 [cs]}}, Jun. 2019, arXiv:
  1906.02243. [Online]. Available: \url{http://arxiv.org/abs/1906.02243}
\BIBentrySTDinterwordspacing

\bibitem{gupta_chasing_2022}
\BIBentryALTinterwordspacing
U.~Gupta, Y.~G. Kim, S.~Lee, J.~Tse, H.-H.~S. Lee, G.-Y. Wei, D.~Brooks, and
  C.-J. Wu, ``Chasing {Carbon}: {The} {Elusive} {Environmental} {Footprint} of
  {Computing},'' \emph{IEEE Micro}, vol.~42, no.~4, pp. 37--47, Jul. 2022.
  [Online]. Available: \url{https://doi.org/10.1109/MM.2022.3163226}
\BIBentrySTDinterwordspacing

\bibitem{gupta2020secure}
A.~Gupta, C.~Lanteigne, and S.~Kingsley, ``Secure: A social and environmental
  certificate for ai systems,'' \emph{arXiv preprint arXiv:2006.06217}, 2020.

\bibitem{ligozat2022unraveling}
A.-L. Ligozat, J.~Lefevre, A.~Bugeau, and J.~Combaz, ``Unraveling the hidden
  environmental impacts of ai solutions for environment life cycle assessment
  of ai solutions,'' \emph{Sustainability}, vol.~14, no.~9, p. 5172, 2022.

\bibitem{kaack_aligning_2021}
\BIBentryALTinterwordspacing
L.~H. Kaack, P.~L. Donti, E.~Strubell, G.~Kamiya, F.~Creutzig, and D.~Rolnick,
  ``Aligning artificial intelligence with climate change mitigation,'' Oct.
  2021. [Online]. Available:
  \url{https://hal.archives-ouvertes.fr/hal-03368037}
\BIBentrySTDinterwordspacing

\bibitem{lannelongue2023carbon}
L.~Lannelongue and M.~Inouye, ``Carbon footprint estimation for computational
  research,'' \emph{Nature Reviews Methods Primers}, vol.~3, no.~1, p.~9, 2023.

\bibitem{bannour2021evaluating}
N.~Bannour, S.~Ghannay, A.~N{\'e}v{\'e}ol, and A.-L. Ligozat, ``Evaluating the
  carbon footprint of nlp methods: a survey and analysis of existing tools,''
  in \emph{Proceedings of the Second Workshop on Simple and Efficient Natural
  Language Processing}, 2021, pp. 11--21.

\bibitem{thompson2020computational}
N.~C. Thompson, K.~Greenewald, K.~Lee, and G.~F. Manso, ``The computational
  limits of deep learning,'' \emph{arXiv preprint arXiv:2007.05558}, 2020.

\bibitem{strubell22}
J.~Dodge, T.~Prewitt, R.~Tachet~des Combes, E.~Odmark, R.~Schwartz,
  E.~Strubell, A.~S. Luccioni, N.~A. Smith, N.~DeCario, and W.~Buchanan,
  ``Measuring the carbon intensity of ai in cloud instances,'' in \emph{ACM
  Conference on Fairness, Accountability, and Transparency}, 2022, p.
  1877–1894.

\bibitem{Henderson20}
P.~Henderson, J.~Hu, J.~Romoff, E.~Brunskill, D.~Jurafsky, and J.~Pineau,
  ``Towards the systematic reporting of the energy and carbon footprints of
  machine learning,'' \emph{Journal of Machine Learning Research}, vol.~21,
  no.~1, 2020.

\bibitem{luccioni2022estimating}
S.~Luccioni, S.~Viguier, and A.-L. Ligozat, ``Estimating the carbon footprint
  of bloom, a 176b parameter language model,'' \emph{arXiv preprint
  arXiv:2211.02001}, 2022.

\bibitem{Arias_IPCC_6}
P.~Arias, N.~Bellouin, E.~Coppola, C.~Jones, G.~Krinner, J.~Marotzke, V.~Naik,
  G.-K. Plattner, M.~Rojas, J.~Sillmann, T.~Storelvmo, P.~Thorne, B.~Trewin,
  K.~Achutarao, B.~Adhikary, K.~Armour, G.~Bala, R.~Barimalala, S.~Berger, and
  K.~Zickfeld, ``Climate change 2021: The physical science basis. contribution
  of working group i to the sixth assessment report of the intergovernmental
  panel on climate change; technical summary.''\hskip 1em plus 0.5em minus
  0.4em\relax IPCC, 2021.

\bibitem{jay:hal-04030223}
\BIBentryALTinterwordspacing
M.~Jay, V.~Ostapenco, L.~Lef{\`e}vre, D.~Trystram, A.-C. Orgerie, and
  B.~Fichel, ``{An experimental comparison of software-based power meters:
  focus on CPU and GPU},'' in \emph{{IEEE/ACM international symposium on
  cluster, cloud and internet computing}}, May 2023. [Online]. Available:
  \url{https://hal.inria.fr/hal-04030223}
\BIBentrySTDinterwordspacing

\bibitem{ligozat:hal-03376391}
\BIBentryALTinterwordspacing
A.-L. Ligozat and S.~Luccioni, ``{A Practical Guide to Quantifying Carbon
  Emissions for Machine Learning researchers and practitioners},'' {MILA ;
  LISN}, Research Report, Jul. 2021. [Online]. Available:
  \url{https://hal.science/hal-03376391}
\BIBentrySTDinterwordspacing

\bibitem{GreenAlgorithms}
\BIBentryALTinterwordspacing
L.~Lannelongue, J.~Grealey, and M.~Inouye, ``Green algorithms: Quantifying the
  carbon emissions of computation,'' \emph{CoRR}, vol. abs/2007.07610, 2020.
  [Online]. Available: \url{https://arxiv.org/abs/2007.07610}
\BIBentrySTDinterwordspacing

\bibitem{CodeCarbon}
\BIBentryALTinterwordspacing
K.~Lottick, S.~Susai, S.~A. Friedler, and J.~P. Wilson, ``Energy usage reports:
  Environmental awareness as part of algorithmic accountability,'' 2019.
  [Online]. Available: \url{https://arxiv.org/abs/1911.08354}
\BIBentrySTDinterwordspacing

\bibitem{Eco2AI}
\BIBentryALTinterwordspacing
S.~Budennyy, V.~Lazarev, N.~Zakharenko, A.~Korovin, O.~Plosskaya, D.~Dimitrov,
  V.~Arkhipkin, I.~Oseledets, I.~Barsola, I.~Egorov, A.~Kosterina, and
  L.~Zhukov, ``Eco2ai: carbon emissions tracking of machine learning models as
  the first step towards sustainable ai,'' 2022. [Online]. Available:
  \url{https://arxiv.org/abs/2208.00406}
\BIBentrySTDinterwordspacing

\bibitem{CarbonTracker}
\BIBentryALTinterwordspacing
L.~F.~W. Anthony, B.~Kanding, and R.~Selvan, ``Carbontracker: Tracking and
  predicting the carbon footprint of training deep learning models,'' 2020.
  [Online]. Available: \url{https://arxiv.org/abs/2007.03051}
\BIBentrySTDinterwordspacing

\bibitem{ExperimentImpactTracker}
P.~Henderson, J.~Hu, J.~Romoff, E.~Brunskill, D.~Jurafsky, and J.~Pineau,
  ``Towards the systematic reporting of the energy and carbon footprints of
  machine learning,'' 2020.

\bibitem{MLCO2}
\BIBentryALTinterwordspacing
A.~Lacoste, S.~Luccioni, V.~Schmidt, and T.~Dandres, ``Quantifying the carbon
  emissions of machine learning,'' 2019. [Online]. Available:
  \url{https://arxiv.org/abs/1910.09700}
\BIBentrySTDinterwordspacing

\bibitem{Cumulator}
M.~J. Tristan~Trebaol, Mary-Anne~Hartley and H.~S. Ghadikolaei, ``A tool to
  quantify and report the carbon footprint of machine learning computations and
  communication in academia and healthcare,'' \emph{Infoscience EPFL: record
  278189}, 2020.

\bibitem{deng2012mnist}
L.~Deng, ``The mnist database of handwritten digit images for machine learning
  research,'' \emph{IEEE Signal Processing Magazine}, vol.~29, no.~6, pp.
  141--142, 2012.

\bibitem{deng2009imagenet}
J.~Deng, W.~Dong, R.~Socher, L.-J. Li, K.~Li, and L.~Fei-Fei, ``Imagenet: A
  large-scale hierarchical image database,'' in \emph{2009 IEEE conference on
  computer vision and pattern recognition}.\hskip 1em plus 0.5em minus
  0.4em\relax Ieee, 2009, pp. 248--255.

\bibitem{maevsky2017evaluating}
D.~Maevsky, E.~Maevskaya, and E.~Stetsuyk, ``Evaluating the ram energy
  consumption at the stage of software development,'' in \emph{Green IT
  Engineering: Concepts, Models, Complex Systems Architectures}.\hskip 1em plus
  0.5em minus 0.4em\relax Springer, 2017, pp. 101--121.

\bibitem{Hodak2019TowardsPE}
M.~Hodak, M.~Gorkovenko, and A.~Dholakia, ``Towards power efficiency in deep
  learning on data center hardware,'' \emph{2019 IEEE International Conference
  on Big Data (Big Data)}, pp. 1814--1820, 2019.

\bibitem{karyakin2017}
A.~Karyakin and K.~Salem, ``A survey of main-memory energy efficiency
  techniques,'' in \emph{Proceedings of the 13th International Workshop on Data
  Management on New Hardware (DaMoN)}.\hskip 1em plus 0.5em minus 0.4em\relax
  Chicago: ACM, 2017, pp. 1--9.

\bibitem{Guo_Yu_2022}
B.~Guo, J.~Yu, D.~Yang, H.~Leng, and B.~Liao, ``Energy-efficient database
  systems: A systematic survey,'' \emph{ACM Computing Surveys}, 2022.

\bibitem{lean-ict}
\BIBentryALTinterwordspacing
{The Shift Project}, ``Lean {ICT}, towards digital sobriety,'' Mar. 2019.
  [Online]. Available: \url{https://theshiftproject.org/en/lean-ict-2/}
\BIBentrySTDinterwordspacing

\bibitem{uptimeinstitute2022}
\BIBentryALTinterwordspacing
U.~Institute, ``2022 data center industry survey,'' 2022. [Online]. Available:
  \url{https://uptimeinstitute.com/uptime_assets/6768eca6a75d792c8eeede827d76de0d0380dee6b5ced20fde45787dd3688bfe-2022-data-center-industry-survey-en.pdf}
\BIBentrySTDinterwordspacing

\bibitem{lawrence2019}
A.~Lawrence, ``{Is PUE Actually Going Up?}'' 2019, [Online; accessed March
  2023].

\bibitem{lawrence2020}
------, ``{Is PUE Actually Going Up?}'' 2020, [Online; accessed March 2023].

\bibitem{ember2022global}
{Ember}, ``Global electricity review 2022,''
  \url{https://ember-climate.org/insights/research/global-electricity-review-2022/},
  2022.

\bibitem{MORO20185}
\BIBentryALTinterwordspacing
A.~Moro and L.~Lonza, ``Electricity carbon intensity in european member states:
  Impacts on ghg emissions of electric vehicles,'' \emph{Transportation
  Research Part D: Transport and Environment}, vol.~64, pp. 5--14, 2018, the
  contribution of electric vehicles to environmental challenges in transport.
  WCTRS conference in summer. [Online]. Available:
  \url{https://www.sciencedirect.com/science/article/pii/S1361920916307933}
\BIBentrySTDinterwordspacing

\bibitem{RYU}
\BIBentryALTinterwordspacing
E.~K. Ryu, J.~Liu, S.~Wang, X.~Chen, Z.~Wang, and W.~Yin, ``Plug-and-play
  methods provably converge with properly trained denoisers,'' 2019. [Online].
  Available: \url{https://arxiv.org/abs/1905.05406}
\BIBentrySTDinterwordspacing

\bibitem{wu_sustainable_2021}
\BIBentryALTinterwordspacing
C.-J. Wu, R.~Raghavendra, U.~Gupta, B.~Acun, N.~Ardalani, K.~Maeng, G.~Chang,
  F.~A. Behram, J.~Huang, C.~Bai, M.~Gschwind, A.~Gupta, M.~Ott, A.~Melnikov,
  S.~Candido, D.~Brooks, G.~Chauhan, B.~Lee, H.-H.~S. Lee, B.~Akyildiz,
  M.~Balandat, J.~Spisak, R.~Jain, M.~Rabbat, and K.~Hazelwood, ``Sustainable
  {AI}: {Environmental} {Implications}, {Challenges} and {Opportunities},''
  \emph{arXiv:2111.00364 [cs]}, Oct. 2021, arXiv: 2111.00364. [Online].
  Available: \url{http://arxiv.org/abs/2111.00364}
\BIBentrySTDinterwordspacing

\bibitem{malmodin2016energy}
J.~Malmodin and D.~Lund{\'e}n, ``The energy and carbon footprint of the ict and
  e\&m sector in sweden 1990-2015 and beyond,'' in \emph{ICT for Sustainability
  2016}.\hskip 1em plus 0.5em minus 0.4em\relax Atlantis Press, 2016, pp.
  209--218.

\bibitem{groger2021green}
J.~Gr{\"o}ger, R.~Liu, L.~Stobbe, J.~Druschke, and N.~Richter, ``Green cloud
  computing,'' \emph{Life cyclebased data collection on environmental impacts
  of cloud computing}, 2021.

\bibitem{guo2022energy}
B.~Guo, J.~Yu, D.~Yang, H.~Leng, and B.~Liao, ``Energy-efficient database
  systems: A systematic survey,'' \emph{ACM Computing Surveys}, vol.~55, no.~6,
  pp. 1--53, 2022.

\bibitem{lefevre2023environmental}
L.~Lef{\`e}vre, A.-L. Ligozat, D.~Trystram, S.~Bouveret, A.~Bugeau, J.~Combaz,
  F.~Emmanuelle, G.~Guennebaud, J.~Lef{\`e}vre, J.-P. Nicola{\"\i}
  \emph{et~al.}, ``Environmental assessment of projects involving ai methods,''
  2023.

\bibitem{WILSON2022101926}
\BIBentryALTinterwordspacing
C.~Wilson and M.~{van der Velden}, ``Sustainable ai: An integrated model to
  guide public sector decision-making,'' \emph{Technology in Society}, vol.~68,
  p. 101926, 2022. [Online]. Available:
  \url{https://www.sciencedirect.com/science/article/pii/S0160791X22000677}
\BIBentrySTDinterwordspacing

\bibitem{scaphandre}
B.~Petit, ``scaphandre,'' 2021.

\end{thebibliography}

\appendix

\section{Methodologies to estimate energy consumption of CPUs and GPUs}\label{AppTDP}
This appendix described the two methods used to estimate energy consumption of CPUs and GPUs

Knowing the model of the CPU or GPU, the first method multiplies the TDP provided by the manufacturer by the duration of training to obtain the energy used in kWh.
TDP is a specification that indicates the maximum amount of power that a computer processor (CPU or GPU) can dissipate when operating at its maximum performance. It refers to the power consumption under the maximum theoretical load. In general, CPUs with a higher number of cores will have a higher TDP because they require more power to operate at maximum performance. 
However, the relationship between TDP and the number of cores is not always straightforward. 
Some CPUs may have a higher TDP even though they have fewer cores, because they are designed to operate at a higher clock speed or have a less efficient architecture.

The second method uses the Intel RAPL (Running Average Power Limit) system management interface integrated in INTEL CPUs or the Power Gadget tool. RAPL allows software to monitor and control the power usage of the processor and its components, such as the CPU cores, memory controllers and GPUs. The Linux powercap driver has the ability to expose the RAPL hardware energy counters by a set of files that can be accessed through the Linux file system. These files make it possible to read the current power usage of the processor and its components, as well as to set power limits to control power usage. Drivers are being developed to get the information from RAPL interface from Windows. A recent implementation is the windows-rapl-driver \footnote{https://github.com/hubblo-org/windows-rapl-driver} from the Scaphandre project \cite{scaphandre}. 

Power Gadget is a standalone software application developed by Intel that provides real-time monitoring of the power usage of Intel processors. It does not rely on the RAPL files, but rather uses its own proprietary methods to access and analyze power consumption data. Power Gadget presents power consumption data in a user-friendly graphical interface that displays real-time power usage of the processor, CPU cores, memory controller, and other components. This tool can be used on Windows and macOS. 

\section{Bugs fix of some software tools }\label{bug}

Some tools must be modified to be used, as they have bugs that have not been fixed by the authors. Here are the changes to make for each one.

\subsection{Experiment-Impact-Tracker}

\begin{itemize}
    \item PyPi package is not the latest, and does not not correspond to documentation (\href{https://github.com/Breakend/experiment-impact-tracker/issues/76}{issue}).
    \item getiterator in file \texttt{/gpu/nvidia.py} must be changed to iter.
    \item For long runs, the INFO log level is too heavy. Change it to the ERROR level.
    \item If you have other experiment-impact-tracker logs in the same folder or subfolders, correct the \texttt{data\_interface.py} file so that the results are shown only from the logs folder that was determined.
\end{itemize}

\subsection{Cumulator}
Correct imports in \texttt{base.py} (structure defined in this file does not correspond to the structure of the package. \href{https://github.com/epfl-iglobalhealth/cumulator/issues/25}{issue})

\subsection{CarbonTracker}
Correct decode deprecated function in Python 3.10 in file \textit{carbontracker/components/gpu/nvidia.py}.

\section{Neural network architectures of experiments}\label{Architectures}

\begin{figure*}[htbp]
\centering
\includegraphics[width=0.7\textwidth]{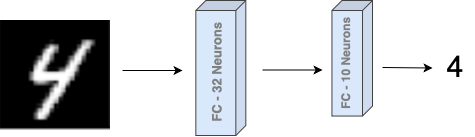}
\caption{Experiment 1 Network Architecture.}
\label{fig:FC}
\end{figure*}

The neural network architecture of Experiment 1 is a fully connected network with a single hidden layer of 32 neurons and an output layer of 10 neurons. The image \ref{fig:FC} shows the architecture.

\begin{figure*}[htbp]
\centering
\includegraphics[width=0.9\textwidth]{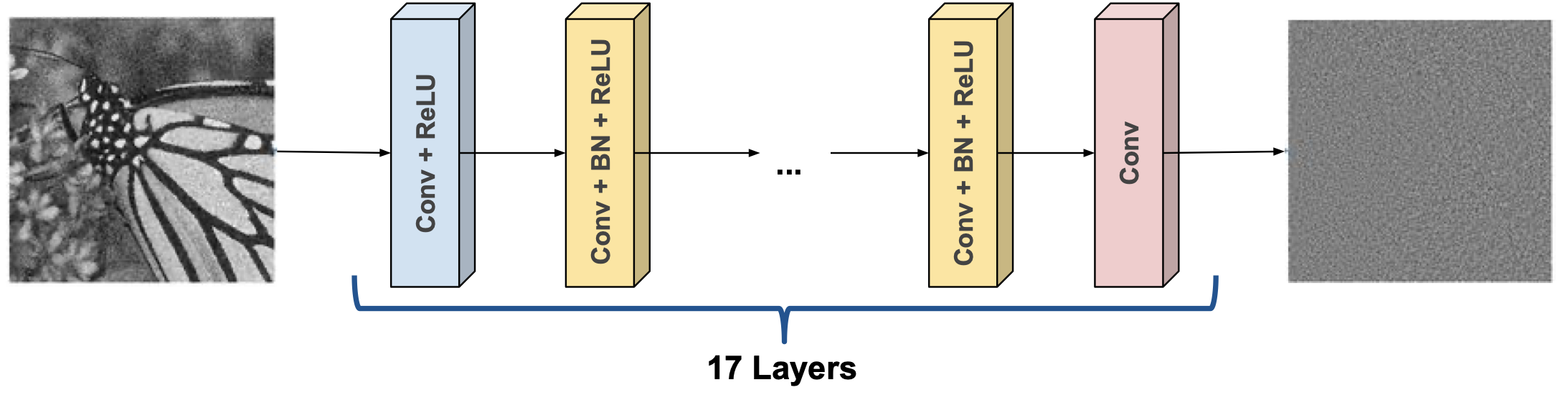}
\caption{DnCNN Network Architecture. Image taken from \cite{RYU}.}
\label{fig:DnCNN}
\end{figure*}

The neural network architecture of Experiment 2 is the DnCNN network presented in \cite{RYU}. The image \ref{fig:DnCNN} shows the architecture proposed in the original paper, which is the one we used in the experiment.

\end{document}